\documentclass[10pt,twocolumn,letterpaper]{article}

\usepackage[pagenumbers]{cvpr} % To force page numbers, e.g. for an arXiv version

\renewcommand{\paragraph}[1]{\vspace{.8em}\noindent\textbf{#1}}

\definecolor{cvprblue}{rgb}{0.21,0.49,0.74}
\usepackage[pagebackref,breaklinks,colorlinks,allcolors=cvprblue]{hyperref}
\usepackage[utf8]{inputenc}
\usepackage[T1]{fontenc}

\usepackage{amsmath}
\usepackage{amsfonts}
\usepackage{amssymb}
\usepackage{nicefrac}
\usepackage{pifont}            % for \ding (check/cross marks)

\usepackage[dvipsnames, table]{xcolor}
\usepackage{graphicx}
\usepackage{rotating}

\usepackage{array}
\usepackage{booktabs}
\usepackage{multirow}
\usepackage{caption}
\usepackage{subcaption}
\usepackage{wrapfig}
\usepackage{microtype}
\usepackage{url}
\usepackage{algorithm}
\usepackage{algpseudocode}

\newcolumntype{x}[1]{>{\centering\arraybackslash}p{#1}}
\newcommand{\tablestyle}[2]{\setlength{\tabcolsep}{#1}\renewcommand{\arraystretch}{#2}\centering\fontsize{8pt}{9pt}\selectfont}
\definecolor{dark_red}{HTML}{d62728}
\newcommand{\authorskip}{\hspace{2.5mm}}
\newcommand{\institutionskip}{\hspace{5.0mm}}

\usepackage{xspace}
\newcommand{\method}{Lacuna\xspace} % model name: change here only
\newcommand{\cmark}{\textcolor{green!60!black}{\ding{51}}}
\newcommand{\xmark}{\textcolor{red!70!black}{\ding{55}}}

\def\confName{CVPR}
\def\confYear{2027}

\title{What Remains and What Is Missing? Learning Point Cloud Semantic Representations for Partial Observations}

\author{Chun-Peng Chang\textsuperscript{\mdseries1} \authorskip 
Shaoxiang Wang\textsuperscript{\mdseries2} \authorskip
Alain Pagani\textsuperscript{\mdseries3} \authorskip
Dariu Gavrila\textsuperscript{\mdseries1} \authorskip
Holger Caesar\textsuperscript{\mdseries1} \authorskip \\[2pt]
\normalsize
\textsuperscript{1}TU Delft, Cognitive Robotics \institutionskip
\textsuperscript{2}RPTU \institutionskip
\textsuperscript{3}DFKI, Augmented Vision
}

\makeatletter
\apptocmd\@maketitle{{\myfigure{}\par}}{}{}
\patchcmd\@maketitle{\vskip .375in}{\vskip 22pt}{}{\PackageError{main}{title skip patch failed}{}}
\patchcmd\@maketitle{\vspace*{24pt}}{\vspace*{14pt}}{}{\PackageError{main}{title-author skip patch failed}{}}
\makeatother

\begin{document}
\newcommand{\myfigure}{
\captionsetup{skip=3pt}
    \centering
    \vspace{-12pt}
    \includegraphics[width=1\linewidth]{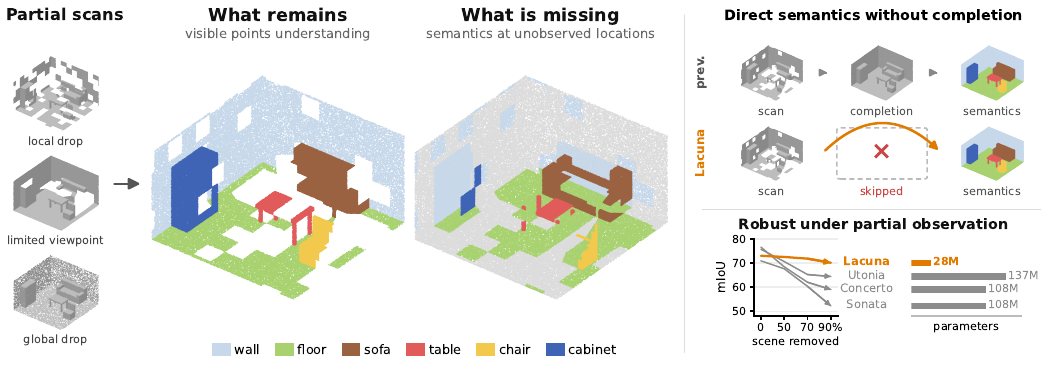}
    \captionsetup{type=figure}\caption{Real scans observed by robot agents are often not complete. Given such a scan, \method produces  semantic features for the points that remain and predicts semantic features at locations that were missing. Unlike the conventional route of  completing geometry before segmenting, we bypass the explicit completion step, which is time-consuming and does not yet perform well. \method shows more robust segmentation performance under partial observation, with less training data and a smaller model than state-of-the-art methods.}
   \label{fig:teaser}
   \vspace{8pt}
}
\maketitle

\begin{abstract}
Self-supervised learning (SSL) yields strong 3D point cloud representations, yet models are primarily evaluated on complete scenes. Robotic agents, by contrast, observe partial scenes due to limited viewpoints and occlusions, raising two questions: can these models (1) understand visible regions in partial inputs, and (2) infer what lies where points are missing without costly 3D completion? Because state-of-the-art 3D SSL methods rely on masked pretraining, one might expect them to handle incomplete point clouds naturally. Surprisingly, performance degrades sharply once points are absent. We show this stems from pretraining design rather than model or data scale: mask tokens reveal the geometry they should hide, compounded by suboptimal backbones and learning frameworks. We propose \textbf{\method}, an SSL point cloud pretraining framework that addresses these issues by removing masked points and predicting their latent features directly from visible context. Using up to $4.8\times$ fewer parameters and less pretraining data, \method{} remains far more robust under missing points and predicts dense semantic features in unobserved regions without explicit completion. \href{https://birdy666.github.io/projects/lacuna/}{Project page}.\vspace*{-12pt}
\end{abstract}

\section{Introduction}
\label{sec:intro}

Existing 3D datasets and benchmarks~\cite{dai2017scannet, yeshwanth2023scannet++, armeni2017joint, caesar2020nuscenes} provide high-quality point clouds that are carefully filtered and capture rich scene detail. This, however, rarely reflects what a robotic agent encounters in practice: its observations are often partial, captured from limited viewpoints, degraded by occlusion, and missing points due to sensor failure  (\cref{fig:teaser}). This gap raises the two challenges addressed in this work.\\
(i) Understanding what remains: whether a model can still produce accurate semantic representations for observed regions under severe data sparsity or extensive missing regions.
(ii) Predicting what is missing: inferring semantic features at unobserved locations conditioned on the visible points.

Recently, self-supervised learning (SSL) approaches for
3D point clouds~\cite{pang2022masked, abdelsamad2026distilling, yang2026towards, abdelsamad2026ghostpoint, yamada20263d, wu2025sonata, zhang2025concerto, zhang2026utonia} have attracted
growing attention. Without relying on expensive human
annotations, these models learn rich semantic features which 
can then be adapted to downstream tasks. At first glance,
the training objective of conventional masking pretext tasks 
seems well aligned with the challenges mentioned above: 
the model is trained to understand a scene from which parts 
have been masked.

\begin{figure*}[t] % 't' helps placement at the top of the page
\captionsetup{skip=3pt}
   \centering
   \includegraphics[width=1\linewidth]{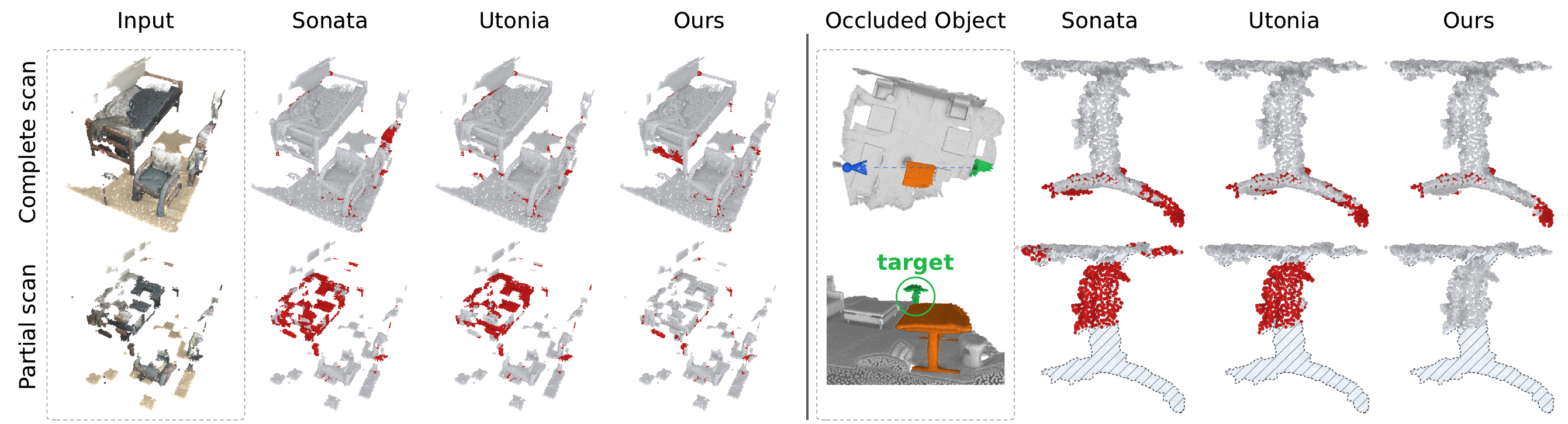}
   \caption{\textbf{Segmentation of partially observed scenes.} Errors of
segmentation on ScanNet val, shown in
\textcolor[rgb]{0.85,0.12,0.12}{red}.
All models do well on the full scan.
\emph{Left:} points removed in small cubic cells, closely aligning with the
masking used in self-supervised pretraining; even so, existing methods break
down once the points are missing.
\emph{Right:} another common source of missing points, occlusion from a
limited viewpoint (camera in \textcolor[rgb]{0.10,0.35,0.95}{blue},
target in \textcolor[rgb]{0.12,0.72,0.28}{green},
occluder in \textcolor[rgb]{0.88,0.38,0.05}{orange}; hatched: the part of the target hidden from the camera).}
   \label{fig:qualitative}
   \vspace{-5mm}
\end{figure*}

Surprisingly, our study suggests otherwise. These models perform well on complete inputs, but their performance degrades sharply when the input point cloud is not complete (\cref{fig:qualitative}), even at the largest scales of data and model size, suggesting that scale alone does not close this gap. We trace it to a shortcut in pretraining. State-of-the-art methods~\cite{wu2025sonata,zhang2025concerto,zhang2026utonia} replace each masked point with a mask token at its original coordinates. Unlike 2D images, whose pixels lie on a fixed grid so that color values carry all the information, in 3D point clouds the mere presence of a point at a given location already reveals a great deal about the scene's geometry. Existing work~\cite{abdelsamad2026distilling} observed that this leakage degrades performance on complete scans. We find a more fundamental problem: the model can minimize the pretext objective without learning features that generalize to scenes where points are actually missing.

To understand what remains, we pretrain on inputs in which masked regions are removed entirely, leaving no geometric trace. Removal alone, however, is not enough. First, we train the encoder to infer the latent features of removed regions through a capacity-limited predictor, forcing the encoder itself to encode missing context, and we align the predictions with the encoder's own feature space. Second, the backbone matters. Most existing methods~\cite{abdelsamad2026distilling, wu2025sonata, zhang2025concerto,zhang2026utonia, chang2026invaria} build on Point Transformer v3~\cite{wu2024point}, which relies on custom modules and scales poorly. While Volt~\cite{yilmaz2026volt} shows that a vanilla Transformer scales better under supervised training, its single-scale design creates a bottleneck under self-supervision: dense self-distillation requires both fine voxels and high-dimensional features, yet global attention scales quadratically with token count. We therefore adopt Volt's simple design but reintroduce a multi-scale hierarchy with windowed attention, achieving both high spatial resolution and wide feature channels. With up to $4.8\times$ fewer parameters and lower latency, the resulting model stays competitive on complete scans. More importantly, it retains higher performance when random regions are removed as in pretraining and generalizes to unseen missing-point patterns, including single-view occlusions, whole-scene corruptions, and sparse outdoor scans.

To predict what is missing, the intuitive route is explicit completion: first reconstruct the missing geometry, then interpret the completed scene. Such completers are diffusion- or flow-based generative models, far larger and orders of magnitude slower than our entire model. Moreover, what a robot needs to know about an unseen region is \emph{what} is there, not its exact appearance, and reconstructing it first solves a harder problem whose errors spread into the observed regions. We instead keep the predictor after pretraining and query it at unobserved locations. It predicts the semantics of missing regions directly and outperforms completion-based pipelines. Although never trained for it, the predicted features also reveal whether an unobserved location is occupied.

In summary, we propose \method, an SSL framework for point clouds under partial observation. (1)~We show that masked pretraining does not naturally make current 3D models robust to missing points, and trace this to mask tokens that leak the geometry of the missing region, compounded by suboptimal backbones and frameworks. (2)~Our framework avoids this shortcut and is far more robust under partial observation, with less training data and a much smaller model. (3)~\method{} predicts dense semantic features at unobserved locations better than completion-based pipelines, without explicit completion, which is costly in both time and model size.

\section{Related Work}
\label{sec:related_work}

\noindent{\bf Point Cloud Backbone. }
Point cloud backbones moved from point-wise MLPs with symmetric pooling~\cite{qi2017pointnet,qi2017pointnet++} to attention over local neighborhoods~\cite{zhao2021point,wu2022point}. To scale to full scenes, PTv3~\cite{wu2024point}, LitePT~\cite{yuelitept2026} and OctFormer~\cite{wang2023octformer} serialize points along space-filling curves and attend within windows of the resulting sequence; Volt~\cite{yilmaz2026volt} shows that a vanilla Transformer over regular volumetric tokens matches these designs without point-specific heuristics.

\noindent{\bf Self-Supervised Learning on Point Clouds. }
Early 3D SSL used geometric pretext tasks and contrastive objectives~\cite{sauder2019self,rao2020global,wang2021unsupervised,xie2020pointcontrast,huang2021spatio,zhang2021self,afham2022crosspoint}. Masked modeling then became dominant: Point-MAE~\cite{pang2022masked} reconstructs masked patches of single objects, MSM~\cite{hermosilla2025masked} predicts masked features of scenes with a decoder, and hybrids add contrastive terms~\cite{qi2023contrast}.  Methods such as Sonata~\cite{wu2025sonata}, Concerto~\cite{zhang2025concerto} and Utonia~\cite{zhang2026utonia} pair masked self-distillation with ever larger models and data. DOS~\cite{abdelsamad2026distilling} removes the mask tokens to address positional leakage for the whole scene segmentation. All of these are evaluated on complete scans. Whether the masked pretext transfers to inputs where points are missing has not been examined, nor has dense semantic prediction at unobserved locations been evaluated. 

\noindent{\bf Robustness and Completion. }
Corruption benchmarks document how point cloud models degrade under dropout and occlusion at object level~\cite{sun2022benchmarking,ren2022benchmarking}, for LiDAR~\cite{kong2023robo3d}, and for indoor segmentation~\cite{raman2024belhouse3d}, without a pretraining remedy. Completion methods instead reconstruct the missing geometry explicitly~\cite{yuan2018pcn,yu2021pointr,song2017semantic,wang2021unsupervised}. Recent work moves this inference into latent space: OccFeat~\cite{sirko2024occfeat} predicts occupancy and features for camera-based BEV networks, GhostPoint~\cite{abdelsamad2026ghostpoint} hallucinates features around detected LiDAR instances, scored by detection; SR-JEPA~\cite{zhou2026sr} imputes one latent for a removed object, scored by its decoded identity. None of these evaluate dense prediction at arbitrary unobserved locations or the features of the points that remain. We address both tasks jointly: robust features for the points that remain, and dense semantic prediction at any unobserved location, without reconstruction.

\section{Semantics Under Partial Observation}
\label{sec:motivation}

What a robot agent sees is rarely the whole scene. Objects hide what stands behind them, the sensor drops returns on dark or glossy surfaces, and a single viewpoint never sees the back of anything (\cref{fig:qualitative}). A model deployed on such data must still recognize the visible points, even though their surroundings are partly gone. And it should be able to say what is at a location it did not see based on the visible cues without explicitly filling in the geometry. 
Let $\mathcal{P}$ denote a complete scene and $\mathcal{P}_v \subset \mathcal{P}$ the points a sensor
actually returns; the rest, $\mathcal{P}_m = \mathcal{P} \setminus \mathcal{P}_v$, is \emph{unobserved}: occluded, dropped by the sensor, or never scanned. Whatever its shape, it leaves no trace in the input: no mask token, no coordinate, nothing that marks where $\mathcal{P}_m$ was.
Let $f_\theta$ be an encoder that maps a point cloud to one feature per input point,
$[f_\theta(\mathcal{P}_v)]_i \in \mathbb{R}^d$ for $\mathbf{p}_i \in \mathcal{P}_v$, and let $h$ be a
semantic head (a linear probe in our experiments) that maps a feature to a class label $y$. 
Under this condition we study two challenges: \\
\textbf{(i) Understanding what remains.} The semantics of every visible point should be recovered
from the partial scan alone, although part of its context is gone:
\begin{equation}
\label{eq:task_remain}
    h\big([f_\theta(\mathcal{P}_v)]_i\big) = y(\mathbf{p}_i), \qquad \forall\, \mathbf{p}_i \in \mathcal{P}_v .
\end{equation}
\textbf{(ii) Predicting what is missing.} For a query location $\mathbf{p}_q \in \mathcal{P}_m$ that was
never observed, a predictor $g$ should return the semantic feature that location would have carried,
given only the query coordinate and the features of the visible scene:
\begin{equation}
\label{eq:task_missing}
    h\big(g(\mathbf{p}_q,\, f_\theta(\mathcal{P}_v))\big) = y(\mathbf{p}_q), \qquad \forall\, \mathbf{p}_q \in \mathcal{P}_m .
\end{equation}

\begin{figure}[t]
  \centering
  \captionsetup{skip=3pt}
   \includegraphics[width=1\linewidth]{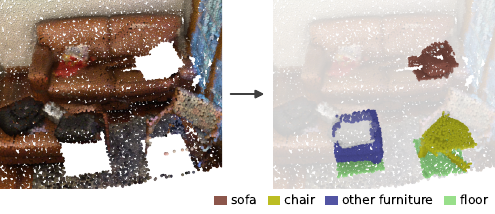}
   \caption{\textbf{Missing area semantic feature prediction.} Given a scene with missing regions (left), predict the semantics of the missing points directly without explicit reconstruction (right).}
   \label{fig:amodal}
    \vspace{-5mm}
\end{figure}

\subsection{Why Masked Pretraining Falls Short}
\label{sec:sfc}
Self-supervised pretraining on masked scenes appears well suited to this task, since its pretext task hides part of a scene and asks the model to understand the rest. Surprisingly, these models often fail as well. We trace this to two causes.

\noindent{\bf An optimization shortcut that leaks geometry.} Replacing masked points with mask tokens at their original positions creates a shortcut during training.
Existing work~\cite{abdelsamad2026distilling} shows that this positional leakage can degrade performance on complete scans. We further show that it creates a bigger problem, one that damages generalization: the pretext task can be solved from the leaked geometry, so the model never has to learn a representation for partial observation.

\noindent{\bf Beyond scaling.} Following the success of large language models~\cite{kaplan2020scaling,hoffmann2022training}, 3D foundation models keep growing in data and model size~\cite{wu2025sonata,zhang2025concerto,zhang2026utonia}. This comes at a cost: larger models are harder to deploy, especially on a robot, high-quality 3D data is scarce, and each step up in scale buys less than the last (Supplement \cref{sec:additional_exp}). We therefore ask whether an efficient framework can achieve robustness to missing points without scaling data or model size, by making better use of the supervision a masked scene already provides.

\begin{figure}[t]
\centering
\captionsetup{skip=5pt}
\includegraphics[width=0.9\linewidth]{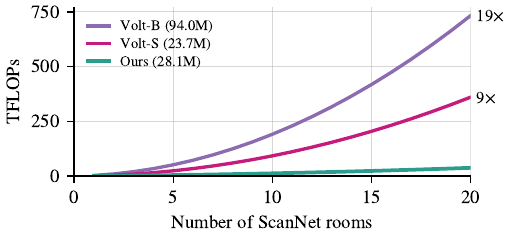}
\caption{\textbf{Compute as scenes grow.} }
\label{fig:volt_scaling}
\vspace{-5mm}
\end{figure}

\noindent{\bf Backbone efficiency.} Removing the shortcut is necessary but not sufficient. Current backbones rely on point-cloud-specific modules~\cite{wang2023octformer,wu2024point,spconv2022} that make them complex to optimize and deploy. Volt~\cite{yilmaz2026volt} shows that a plain Transformer of standard operations can replace them, but at a single scale it pays two costs as scenes grow and features widen. First, every non-empty voxel of size $v$ becomes a token, and since points sample surfaces, a scene of surface area $A$ yields $N \approx A/v^{2}$ tokens. With global attention and a feature dimension $d$ shared by all $L$ layers, the encoder costs
\begin{equation}
    \mathcal{O}\!\left(L\left(N d^{2} + N^{2} d\right)\right) = \mathcal{O}\!\left(L\left(\frac{A}{v^{2}}\, d^{2} + \frac{A^{2}}{v^{4}}\, d\right)\right).
\end{equation}
As scenes grow, keeping the cost fixed requires $v \propto \sqrt{A}$, which loses fine geometry, while keeping $v$ fixed makes the cost grow quadratically with $A$ (\cref{fig:volt_scaling}).
Second, because $d$ is shared, raising it by $\alpha$ for the deep layers, which need high-dimensional features, raises the cost of \emph{every} layer by up to $\alpha^{2}$, including early layers that need far less capacity.

\subsection{Why Explicit Completion Falls Short}
\label{sec:dense_pred}
An intuitive route to task~(ii) is explicit completion: a generative model~\cite{meng2026seen2scene} first reconstructs the missing geometry $\hat{\mathcal{P}}_m$, and the encoder then reads the semantics at $\mathbf{p}_q$ from $f_\theta(\mathcal{P}_v \cup \hat{\mathcal{P}}_m)$. This route has three drawbacks. First, it is expensive: a recent scene completer~\cite{meng2026seen2scene} alone has over $20\times$ the parameters of our entire model and takes over $20$ minutes per ScanNet scene on an A100 (\cref{tab:explicit}). Second, the surfaces it generates when applied zero-shot are not realistic enough to be encoded directly: filling the holes with them sometimes performs worse than leaving the holes empty, for every encoder we test, and degrades even the visible points (\cref{sec:additional_exp}). Third, it solves a harder problem than needed: a robot acting in a scene needs to know what is in the missing regions, not to reproduce their exact surfaces. We therefore predict the semantics at $\mathbf{p}_q$ directly from $\mathcal{P}_v$, without reconstructing any geometry (\cref{sec:dense_inference}).

\begin{figure}[t]
  \centering
  \captionsetup{skip=5pt}
   \includegraphics[width=0.9\linewidth]{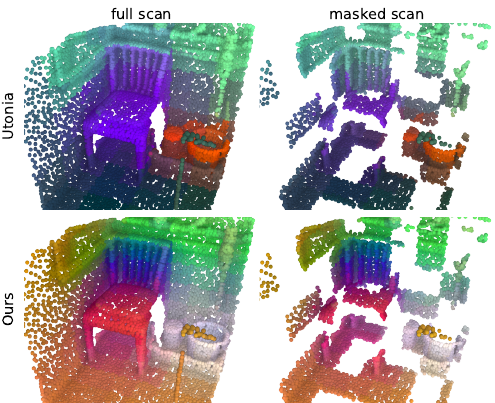}
   \caption{\textbf{PCA of frozen features}, full scan vs.\ masked scan. Once points are missing, existing work often produces inconsistent features (e.g., the chair back blends into the background).}
   \label{fig:feat_pca}
   \vspace{-2mm}
\end{figure}

\begin{figure*}[t]
  \centering
  \captionsetup{skip=5pt}
   \includegraphics[width=1\linewidth]{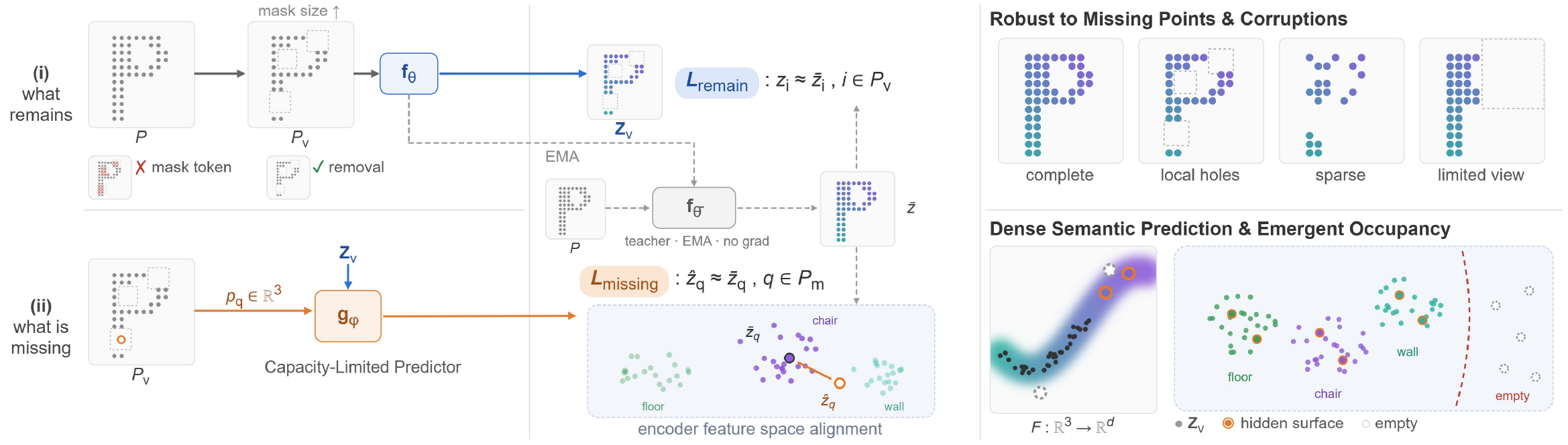}
   \caption{\textbf{Overview of \method.}
\textbf{(i)~What remains:} the student $f_\theta$ sees only the visible points $P_v$ (masked points are removed, not replaced by tokens that leak their geometry); $\mathcal{L}_{\text{remain}}$ matches $Z_v$ to an EMA teacher on the complete scene.
\textbf{(ii)~What is missing:} a capacity-limited predictor $g_\phi$ reads the feature $\hat{z}_q$ of any query $p_q$ from $Z_v$; $\mathcal{L}_{\text{missing}}$ aligns it with the teacher's in the encoder's own feature space.
\textbf{Right:} features stay stable as points go missing, and $f_\theta$ with $g_\phi$ forms a feature field in which hidden surfaces fall within their class and empty space stays linearly separable, although never queried in training.}
   \label{fig:framework}
   \vspace{-4mm}
\end{figure*}

\section{Method}

\subsection{Backbone Architecture}
\label{sec:backbone}
\noindent{\bf Allocating Resolution and Capacity with a Hierarchy.}
We follow the plain Transformer design~\cite{yilmaz2026volt} and add one ingredient, a hierarchy over resolution and width, to remove the two costs of a single-scale design (\cref{sec:sfc}). A hierarchy spends resolution and width where each pays off. Early levels keep fine voxels at a small width and attend within local windows, so object geometry is preserved at a cost that grows linearly with the scene. Windows are fixed cubes of $16^3$ cells on the voxel grid, holding whatever tokens fall inside, and alternate blocks shift the lattice by half a window so that information crosses window borders~\cite{liu2021swin}. Later levels pool to coarser tokens and widen the features where semantics form. The coarsest level attends globally, which a room-scale scene affords; for larger scenes it can be windowed like the others (details in \cref{sec:model_details} \& \cref{sec:why_hierarchy}). A normalization follows every pooling projection; without it, feature magnitudes grow unchecked over pretraining.

\subsection{Self-Supervised Pretraining Framework}
\label{sec:ssl}
\cref{fig:framework} gives an overview of our pretraining, and \cref{fig:feat_pca} the features it yields under missing points.

\noindent{\bf Bootstrapping from What Remains to What Is Missing.}
We train the two tasks of \cref{sec:sfc} directly. Given a scene $\mathcal{P} = \{\mathbf{p}_i\}_{i=1}^{N}$, we sample masked regions and split it into visible points $\mathcal{P}_v$ and masked points $\mathcal{P}_m = \mathcal{P} \setminus \mathcal{P}_v$. A teacher $f_{\bar\theta}$ encodes the complete scene and provides target features $\bar{\mathbf{z}}$; the student $f_\theta$ encodes only $\mathcal{P}_v$. The training loss is
\begin{equation}
    \mathcal{L} =
    \underbrace{\frac{1}{|\mathcal{P}_v|}\sum_{i \in \mathcal{P}_v} \ell_{\text{vis}}\big(\mathbf{z}_i, \bar{\mathbf{z}}_i\big)}_{\mathcal{L}_{\text{remain}}}
    \; + \; w(t)\,
    \underbrace{\frac{1}{|\mathcal{P}_m|}\sum_{q \in \mathcal{P}_m} \ell_{\text{mask}}\big(\hat{\mathbf{z}}_q, \bar{\mathbf{z}}_q\big)}_{\mathcal{L}_{\text{missing}}},
\end{equation}
where $\ell_{\text{vis}}$ is a prototype assignment loss~\cite{caron2021emerging, wu2025sonata} that constrains the encoder at every visible point, without a predictor in between, and $\ell_{\text{mask}}$ supervises a prediction $\hat{\mathbf{z}}_q$ at every removed point. $\mathcal{L}_{\text{remain}}$: from the partial scan alone, the student must match what the teacher sees in the complete one. $\mathcal{L}_{\text{missing}}$: the feature of a point that was never observed must be inferred from the visible scene. Both terms follow a schedule: the mask size $b(t)$ and ratio $r(t)$ grow early in training, so $\mathcal{P}_v$ shrinks while $\mathcal{P}_m$ grows, and the weight $w(t)$ ramps from $0$ to $1$. Training therefore starts as self-distillation on visible points and progressively hands the signal to missing point prediction, from local geometry to object- and scene-level context.

\noindent{\bf Masking by Removal.}
Mask tokens would keep $\mathcal{P}_m$ in the input as $\mathcal{P}_v \cup \{(\mathbf{p}_j, \mathbf{m})\}_{j \in \mathcal{P}_m}$, and with it the geometry of the missing region (\cref{sec:sfc}). We instead remove the masked points~\cite{pang2022masked,abdelsamad2026distilling}, so the encoder sees only what a sensor would, $\mathbf{Z}_v = f_\theta(\mathcal{P}_v)$, and the missing context enters only through the teacher, as a target rather than as an input.

\noindent{\bf Capacity-Limited Predictor and Feature-Space Alignment.}
We view the encoder output as a continuous \emph{feature field} $\mathcal{F}: \mathbb{R}^3 \rightarrow \mathbb{R}^d$, observed only at the visible points. A predictor $g_\phi$, as in JEPA-style frameworks~\cite{assran2023self, saito2025point}, retrieves its value at a query location $\mathbf{p}_q \in \mathbb{R}^3$, in training a removed point of $\mathcal{P}_m$:
\begin{equation}
    \hat{\mathbf{z}}_q = g_\phi\big(\gamma(\mathbf{p}_q),\, \mathbf{Z}_v\big) \approx \mathcal{F}(\mathbf{p}_q).
\end{equation}
A learned query token with a positional embedding $\gamma(\mathbf{p}_q)$ attends to the visible keys and values $\mathbf{Z}_v$, with 3D rotary encoding of their relative offsets (\cref{sec:model_details}); queries never interact~\cite{fu2024rethinking,leijenaar2026asymmetric}, so each prediction depends only on the query location and the visible scene. Two choices define this predictor. First, it is shallow ($|\phi| \ll |\theta|$), a single cross-attention layer: a predictor with enough capacity infers the missing semantics on its own, whereas a shallow one can only read what the visible features already encode, so the encoder must carry the information about what is missing. Second, its output is held in the encoder's own feature space,
\begin{equation}
    \ell_{\text{mask}}\big(\hat{\mathbf{z}}_q, \bar{\mathbf{z}}_q\big) = \ell_{\text{vis}}\big(\hat{\mathbf{z}}_q, \bar{\mathbf{z}}_q\big) + \lambda\, \big\|\hat{\mathbf{z}}_q - \bar{\mathbf{z}}_q\big\|_{\text{s}\ell_1},
\end{equation}
with features standardized per channel and $\|\cdot\|_{\text{s}\ell_1}$ the smooth-$\ell_1$ distance. The prototype term asks only that a hidden position fall into the right cluster; the alignment term asks for the feature itself. Visible and predicted features thus form one field, and because of this alignment the predictor is not discarded after pretraining but reused at inference to predict the features of unobserved locations.

\subsection{Dense Semantic Prediction}
\label{sec:dense_inference}
Predicting what is missing is more than just an additional task. To predict a removed region, the encoder must encode context it could otherwise ignore, so the prediction task also improves the features at the visible points. Because $\ell_{\text{mask}}$ holds $\hat{\mathbf{z}}_q$ in the encoder's own feature space, a single head $h$ trained on visible points reads the field $\mathcal{F}$ anywhere. After pretraining, the frozen encoder and predictor therefore solve task~(ii), \cref{eq:task_missing}, with $g = g_\phi$: one encoder pass, then one cross-attention layer per query $\mathbf{p}_q \in \mathbb{R}^3$, without reconstructing geometry, in milliseconds. In evaluation, $\mathcal{P}_m$ only decides where the queries are placed; at deployment, the predicted features also separate hidden surfaces from empty space.

% Shading: best (dark, bold) / second (light) among the first five rows; best only within the dagger pair; clean not shaded.
% '#Scenes' column: no legend entry, it wrapped the legend onto a third line.
\begin{table*}[t]
    \centering
    \captionsetup{skip=3pt}
    \caption{\textbf{Robustness to missing points.} ScanNet val mIoU with frozen backbones. The scene is divided into cubic cells of size $b$ (m) and a fraction $r$ ($\%$) of the cells is removed from the input, written $b$@$r$. \textit{Left:} the points that survive removal are biased toward easy classes, so every column, including clean, is scored on a fixed $1\%$ of the points that is never removed. \textit{Right:} removed cells are fully empty; scored on the visible points of objects with over $90\%$ of their points removed. Protocols in \cref{sec:eval_protocol}. \\
    \scriptsize{val/test: val and test scenes included in pretraining; 2D: image-distilled pretraining; Params: deployed feature extractor; Latency: forward latency per ScanNet val scene on one A100. (lin.): linear probe on the frozen backbone. $^{\dagger}$Pretrained on the ScanNet and ScanNet++ train splits only. $^{\S}$Evaluated with MSM's own segmentation head}.}
    \label{tab:probe}
    \footnotesize
    \setlength{\tabcolsep}{2.5pt}
    \begin{tabular}{l | cccc c | c cccccc | cccc}
    \toprule
    \multirow{2}{*}{Model (lin.)} & \multirow{2}{*}{val/test} & \multirow{2}{*}{2D} & \multirow{2}{*}{\#Scenes} & \multirow{2}{*}{Params} & Latency & \multicolumn{7}{c|}{Random mask (fixed $1\%$ scored)} & \multicolumn{4}{c}{Objects $>\!90\%$ hidden} \\
    \cmidrule(lr){7-13}\cmidrule(lr){14-17}
    & & & & & (ms) & \textcolor{gray}{clean} & {\scriptsize 0.2@50} & {\scriptsize 0.4@50} & {\scriptsize 0.2@70} & {\scriptsize 0.4@70} & {\scriptsize 0.2@90} & {\scriptsize 0.4@90} & {\scriptsize 0.2@70} & {\scriptsize 0.4@70} & {\scriptsize 0.2@90} & {\scriptsize 0.4@90} \\
    \midrule
        MSM$^{\S}$~\cite{hermosilla2025masked} & \cmark & \xmark & 1.6k & \textit{67.6M} & 115 & \textcolor{gray}{66.6} & 63.7 & 61.5 & 59.4 & 56.4 & \cellcolor[HTML]{eef9f0}49.6 & \cellcolor[HTML]{eef9f0}46.1 & 34.8 & 44.4 & 47.8 & 44.5 \\
        Sonata~\cite{wu2025sonata} & \cmark & \xmark & 140k & \textit{108.5M} & 86 & \textcolor{gray}{70.9} & 68.1 & 64.4 & 61.1 & 50.0 & 31.6 & 25.9 & 34.8 & 37.1 & 25.3 & 31.7 \\
        Concrt.~\cite{zhang2025concerto} & \cmark & \cmark & 40k & \textit{108.5M} & 84 & \textcolor{gray}{76.7} & \cellcolor[HTML]{def3e4}\bf 72.7 & 67.1 & 61.6 & 53.9 & 38.2 & 29.3 & 31.9 & 44.7 & 41.5 & 45.2 \\
        Utonia~\cite{zhang2026utonia} & \cmark & \cmark & 250k & \textit{137.3M} & 117 & \textcolor{gray}{75.9} & 71.9 & \cellcolor[HTML]{eef9f0}70.7 & \cellcolor[HTML]{eef9f0}62.7 & \cellcolor[HTML]{eef9f0}62.6 & 44.7 & 40.7 & \cellcolor[HTML]{eef9f0}35.1 & \cellcolor[HTML]{eef9f0}50.5 & \cellcolor[HTML]{eef9f0}52.0 & \cellcolor[HTML]{eef9f0}51.9 \\
        \method{} & \cmark & \xmark & 2.6k & \textit{28.1M} & \bf 34 & \textcolor{gray}{73.0} & \cellcolor[HTML]{eef9f0}72.1 & \cellcolor[HTML]{def3e4}\bf 71.2 & \cellcolor[HTML]{def3e4}\bf 70.7 & \cellcolor[HTML]{def3e4}\bf 68.6 & \cellcolor[HTML]{def3e4}\bf 65.4 & \cellcolor[HTML]{def3e4}\bf 61.9 & \cellcolor[HTML]{def3e4}\bf 36.8 & \cellcolor[HTML]{def3e4}\bf 54.6 & \cellcolor[HTML]{def3e4}\bf 59.5 & \cellcolor[HTML]{def3e4}\bf 52.2 \\
    \midrule
        Sonata$^{\dagger}$~\cite{wu2025sonata} & \xmark & \xmark & 2.1k & \textit{41.5M} & 68 & \textcolor{gray}{63.7} & 59.7 & 56.2 & 50.4 & 39.9 & 22.9 & 17.9 & 25.1 & 27.5 & 17.8 & 22.6 \\
        \method{}$^{\dagger}$ & \xmark & \xmark & 2.1k & \textit{28.1M} & \bf 34 & \textcolor{gray}{71.5} & \cellcolor[HTML]{def3e4}\bf 71.2 & \cellcolor[HTML]{def3e4}\bf 69.6 & \cellcolor[HTML]{def3e4}\bf 69.7 & \cellcolor[HTML]{def3e4}\bf 67.1 & \cellcolor[HTML]{def3e4}\bf 64.2 & \cellcolor[HTML]{def3e4}\bf 60.0 & \cellcolor[HTML]{def3e4}\bf 31.8 & \cellcolor[HTML]{def3e4}\bf 51.4 & \cellcolor[HTML]{def3e4}\bf 56.1 & \cellcolor[HTML]{def3e4}\bf 47.4 \\
    \bottomrule
    \end{tabular}
    \vspace{-3mm}
\end{table*}

\begin{table*}[t]
    \centering
    \begin{minipage}[t]{0.48\textwidth}
    \centering
        \centering
    \captionsetup{skip=3pt}
    \caption{
    \textbf{Comparison with explicit completion.} ScanNet val mIoU over the whole scene, visible and removed points (details in \cref{sec:additional_exp}). S2S~\cite{meng2026seen2scene} reconstructs the removed regions before encoding; ours predicts features at the removed positions without explicit reconstruction or completion. 
    \scriptsize{Params: encoder + \textit{S2S}, or encoder + \textit{predictor} for ours}.
    }
    \label{tab:explicit}
    \tablestyle{3pt}{1.08}
    \resizebox{0.76\linewidth}{!}{%
    \begin{tabular}{c l | c | c c | c c | c c}
    \toprule
    & \multirow{2}{*}{Model} & \multirow{2}{*}{Params (M)} & \multicolumn{2}{c|}{$r{=}0.5$} & \multicolumn{2}{c|}{$r{=}0.7$} & \multicolumn{2}{c}{Single View} \\
    \cmidrule(lr){4-5}\cmidrule(lr){6-7}\cmidrule(lr){8-9}
    & & & $b{=}0.2$ & $0.4$\ & $0.2$\ & $0.4$\ & $>\!75\%$ & $>\!90\%$ \\
    \midrule
    \cellcolor[HTML]{D3D3D3} & MSM & 68 + \textit{681} & 26.6 & 29.0 & 15.8 & 16.7 & 29.9 & 32.2 \\
    \cellcolor[HTML]{D3D3D3} & Sonata & 109 + \textit{681} & 43.8 & 42.4 & 26.2 & 22.2 & 36.4 & 39.3 \\
    \cellcolor[HTML]{D3D3D3} & Concrt. & 109 + \textit{681} & 53.6 & 50.0 & 30.2 & 24.4 & 43.3 & 47.7 \\
    \cellcolor[HTML]{D3D3D3}\multirow{-4}{*}{\rotatebox[origin=c]{90}{\shortstack{w/ S2S}}} & Utonia & 137 + \textit{681} & 54.1 & 48.3 & 28.0 & 23.1 & 44.0 & 47.2 \\
    \midrule
    & \method{} & \cellcolor[HTML]{def3e4}29 & \cellcolor[HTML]{def3e4}\bf 67.7 & \cellcolor[HTML]{def3e4}\bf 66.7 & \cellcolor[HTML]{def3e4}\bf 65.2 & \cellcolor[HTML]{def3e4}\bf 63.1 & \cellcolor[HTML]{def3e4}\bf 61.0 & \cellcolor[HTML]{def3e4}\bf 62.8 \\
    \bottomrule
    \end{tabular}}

    \end{minipage}
    \hfill%
    \begin{minipage}[t]{0.48\textwidth}
    \centering
        \centering
    \captionsetup{skip=3pt}
    \caption{\textbf{Cross-domain generalization to outdoor datasets.} All models here are trained on indoor data only; Utonia's pretraining includes both outdoor datasets. We report mIoU on nuScenes val and SemanticKITTI val with a linear head trained on the frozen encoder.}
    \label{tab:transfer}
    \setlength{\tabcolsep}{6pt}
        \footnotesize
        \resizebox{\linewidth}{!}{%
        \begin{tabular}{l ccc | cc}
        \toprule
        Model & Indoor & 2D & ScanNet val & nuScenes & SemanticKITTI \\
        \midrule
        MSM & \cmark & \xmark & \textcolor{gray}{66.3} & 30.3 & 29.7 \\
        Sonata & \cmark & \xmark & \textcolor{gray}{70.9} & 41.8 & 43.5 \\
        Concrt. & \cmark & \cmark & \textcolor{gray}{76.6} & 40.9 & 42.0 \\
        \method{} & \cmark & \xmark & \textcolor{gray}{73.0} & \cellcolor[HTML]{def3e4}\textbf{43.1} & \cellcolor[HTML]{def3e4}\textbf{44.0} \\
        \bottomrule
        \end{tabular}}

    \end{minipage}
    \vspace{-5mm}
\end{table*}

\section{Experiments}
\label{sec:exp}
\noindent{\bf Environment \& Evaluation Protocol.}
All models are evaluated on a single NVIDIA A100 GPU. We report mIoU without test-time augmentation (TTA), since TTA adds inference cost that is often impractical in time-sensitive settings and its strategies vary across prior work, which makes comparisons without it more direct.  Every backbone is frozen and every head is trained on complete scans, so every drop we report is caused by the input alone. We distinguish in-distribution robustness, where points are removed in blocks like the pretraining masks, from zero-shot generalization to conditions no model has seen in training: whole-scene corruptions, single-viewpoint occlusion and sparse outdoor LiDAR.
Many prior works include all available scenes (with val/test split), to maximize the scale of pretraining data. However, it remains debatable whether this setting faithfully evaluates a model's ability to generalize to unseen scenes. For transparency, we report our results with and without validation and test scenes included in pretraining.  
More details can be found in the supplement (\cref{sec:env_data} \& \cref{sec:eval_protocol}).

\begin{table}[t]
    \centering
    \captionsetup{skip=3pt}
    \caption{\textbf{Backbone and Pretraining: }All models are pretrained on the ScanNet and ScanNet++ train splits only.}
    \label{tab:cross_backbone}
    \setlength{\tabcolsep}{4pt}
    \small
    \resizebox{\linewidth}{!}{%
    \begin{tabular}{l r l | c ccc | cc}
    \toprule
    \multirow{2}{*}{Backbone} & \multirow{2}{*}{Params} & \multirow{2}{*}{Pretraining} & \multirow{2}{*}{clean \(\uparrow\)} & \multicolumn{3}{c|}{Random mask, $b{=}0.4$} & \multicolumn{2}{c}{$>\!90\%$ hidden} \\
    & & & & @50 & @70 & @90 & @70 & @90 \\
    \midrule
        PTv3 & 41.5M & Sonata & 63.7 & 56.2 & 39.9 & 17.9 & 27.5 & 22.6 \\ 
        PTv3 & 41.5M & \method{} & 63.7 & 56.9 & 50.2 & 35.8 & 34.7 & 36.4 \\
        Volt-S & 23.7M & Sonata & 61.0 & 57.1 & 46.1 & 33.8 & 36.0 & 29.0 \\
        Volt-S & 23.7M & \method{} & 62.2 & 60.5 & \cellcolor[HTML]{eef9f0}58.6 & \cellcolor[HTML]{eef9f0}54.5 & \cellcolor[HTML]{eef9f0}45.0 & \cellcolor[HTML]{eef9f0}40.3 \\
        Volt-B & 94.0M & \method{} & 63.3 & 61.0 & 58.4 & 53.2 & 44.4 & 37.7 \\
    \method{} & 28.1M & Sonata & \cellcolor[HTML]{eef9f0}68.9 & \cellcolor[HTML]{eef9f0}63.7 & 54.7 & 35.4 & 43.0 & 35.8 \\
        \cellcolor[HTML]{cce3de}\method{} & \cellcolor[HTML]{cce3de}28.1M & \cellcolor[HTML]{cce3de}\method{} & \cellcolor[HTML]{def3e4}\bf 71.5 & \cellcolor[HTML]{def3e4}\bf 69.6 & \cellcolor[HTML]{def3e4}\bf 67.1 & \cellcolor[HTML]{def3e4}\bf 60.0 & \cellcolor[HTML]{def3e4}\bf 51.4 & \cellcolor[HTML]{def3e4}\bf 47.4 \\
    \bottomrule
    \end{tabular}}
    \vspace{-4mm}
\end{table}

\subsection{What Remains?}
\noindent{\bf Does SSL Pretraining Naturally Make Models Robust to Locally Missing Points?}
\cref{tab:probe} evaluates models\footnote{Excluded from comparison because source code is unreleased~\cite{abdelsamad2026distilling,abdelsamad2026ghostpoint,zhou2026sr}.} under the common masking protocol of self-supervised pretraining: the scene is divided into cubic cells of size $b$ and a fraction $r$ of them is removed. Since small and difficult objects vanish first, scoring the surviving points would make the task easier as more is removed. 
We therefore use two protocols. The first scores a fixed 1\% of the points, exempt from removal, so that every setting is compared on the same points. The second leaves the removed cells fully empty and scores the visible points of objects that are more than 90\% hidden (Sec. D).
Surprisingly, the self-supervised models degrade significantly as well, even though the input matches their own pretext task (see \cref{sec:additional_exp} for the results of supervised methods).

\begin{table*}[t]
    \centering
    \captionsetup{skip=3pt}
    \caption{\textbf{Zero-shot generalization: corruptions and a limited viewpoint.} ScanNet val mIoU. \textit{Left:} corruptions follow PointCloud-C~\cite{ren2022benchmarking} protocol, with drop\_global (random fraction of points removed), add\_local (spurious points added in small clusters) and jitter (Gaussian coordinate noise). \textit{Right:} one interior camera per scene; scored on the visible points of objects more than $75\%$ / $90\%$ hidden from that viewpoint (walls and floors excluded), with the full scene (full) or single view (single) as input.}
    \label{tab:corruption}\label{tab:viewpoint}
    \setlength{\tabcolsep}{4pt}
    \footnotesize
    \begin{tabular}{l cc cc cc | cc cc}
    \toprule
    \multirow{2}{*}{Model} & \multicolumn{2}{c}{drop\_global} & \multicolumn{2}{c}{add\_local} & \multicolumn{2}{c|}{jitter} & \multicolumn{2}{c}{$>\!75\%$ hidden} & \multicolumn{2}{c}{$>\!90\%$ hidden} \\
    \cmidrule(lr){2-3}\cmidrule(lr){4-5}\cmidrule(lr){6-7}\cmidrule(lr){8-9}\cmidrule(lr){10-11}
    & 90\% & 99\% & 1\% & 10\% & 1\,cm & 5\,cm & full & single & full & single \\
    \midrule
    MSM$^{\S}$~\cite{hermosilla2025masked} & \cellcolor{Salmon!27}53.6 (-19.2\%) & \cellcolor{Salmon!58}28.3 (-57.3\%) & \cellcolor{Salmon!2}66.1 (-0.5\%) & \cellcolor{Salmon!20}57.9 (-12.7\%) & \cellcolor{Salmon!5}65.3 (-1.6\%) & \cellcolor{Salmon!58}28.3 (-57.3\%) & \textcolor{gray}{52.1} & 49.5 & \textcolor{gray}{51.1} & 38.3 \\
    Sonata~\cite{wu2025sonata} & \cellcolor{Salmon!10}67.4 (-4.9\%) & \cellcolor{Salmon!69}18.9 (-73.3\%) & \cellcolor{Salmon!4}70.2 (-1.2\%) & \cellcolor{Salmon!33}52.5 (-25.9\%) & \cellcolor{Salmon!1}70.8 (-0.3\%) & \cellcolor{Salmon!43}44.0 (-38.0\%) & \textcolor{gray}{60.7} & 53.2 & \textcolor{gray}{56.0} & 36.3 \\
    Concrt.~\cite{zhang2025concerto} & \cellcolor{Salmon!11}72.5 (-5.6\%) & \cellcolor{Salmon!70}11.8 (-84.6\%) & \cellcolor{Salmon!3}76.1 (-0.9\%) & \cellcolor{Salmon!14}70.8 (-7.7\%) & \cellcolor{Salmon!3}76.2 (-0.7\%) & \cellcolor{Salmon!38}52.7 (-31.3\%) & \textcolor{gray}{66.5} & 57.7 & \textcolor{gray}{64.8} & 44.1 \\
    Utonia~\cite{zhang2026utonia} & \cellcolor{Salmon!7}73.6 (-2.8\%) & \cellcolor{Salmon!70}16.5 (-78.2\%) & 75.8 (0.0\%) & \cellcolor{Salmon!11}71.6 (-5.4\%) & 75.8 (0.0\%) & \cellcolor{Salmon!69}20.6 (-72.8\%) & \textcolor{gray}{66.2} & \cellcolor[HTML]{eef9f0} 58.5 & \textcolor{gray}{65.2} & \cellcolor[HTML]{def3e4}\bf 46.7 \\
    \method{} & \cellcolor{Salmon!7}71.2 (-2.5\%) & \cellcolor{Salmon!35}52.5 (-28.2\%) & 73.1 (0.0\%) & \cellcolor{Salmon!7}71.1 (-2.7\%) & 73.4 (+0.5\%) & \cellcolor{Salmon!27}58.7 (-19.7\%) & \textcolor{gray}{65.9} & \cellcolor[HTML]{def3e4}\bf 59.1 & \textcolor{gray}{60.6} & \cellcolor[HTML]{eef9f0} 46.2 \\
    \midrule
    Sonata$^{\dagger}$~\cite{wu2025sonata} & \cellcolor{Salmon!16}57.9 (-9.4\%) & \cellcolor{Salmon!70}7.3 (-88.5\%) & \cellcolor{Salmon!2}63.5 (-0.6\%) & \cellcolor{Salmon!40}42.5 (-33.4\%) & 64.1 (+0.3\%) & \cellcolor{Salmon!58}27.1 (-57.5\%) & \textcolor{gray}{54.2} & 40.9 & \textcolor{gray}{48.6} & 26.9 \\
    % Shade = round(70 min(1, drop/75)^0.7).
    \method{}$^{\dagger}$ & \cellcolor{Salmon!7}69.5 (-2.9\%) & \cellcolor{Salmon!37}50.2 (-29.8\%) & 71.5 (+0.1\%) & \cellcolor{Salmon!13}66.6 (-6.8\%) & 72.4 (+1.3\%) & \cellcolor{Salmon!27}57.5 (-19.6\%) & \textcolor{gray}{61.5} & \cellcolor[HTML]{def3e4}\bf 54.0 & \textcolor{gray}{56.8} & \cellcolor[HTML]{def3e4}\bf  44.4 \\
    \bottomrule
    \end{tabular}
    \vspace{-2mm}
\end{table*}

\begin{table*}[t]
\centering
\captionsetup{skip=3pt}
\caption{\textbf{Semantic segmentation at unobserved positions.} mIoU (ScanNet val) at removed positions. The feature at a removed position is copied from the nearest visible point (NN), averaged over the $k$ nearest ($k$NN), or predicted directly where the model has a mechanism for it. Left: block masks of $0.4$ / $0.8$\,m at ratio $0.5$ / $0.7$. Right: hidden points of objects more than $75\%$ / $90\%$ hidden from the single viewpoint. \scriptsize{$^{\S}$MSM's decoder is queried in chunks of $100$ positions, so it cannot read the missing geometry from the layout of the targets.}}
\label{tab:amodal}\label{tab:amodal-vp}
\setlength{\tabcolsep}{3pt}
\footnotesize
\resizebox{\textwidth}{!}{%
\begin{tabular}{l ccc ccc ccc ccc | ccc ccc}
\toprule
\multirow{2}{*}{Model} & \multicolumn{3}{c}{$0.4$\,m @ $0.5$} & \multicolumn{3}{c}{$0.8$\,m @ $0.5$} & \multicolumn{3}{c}{$0.4$\,m @ $0.7$} & \multicolumn{3}{c|}{$0.8$\,m @ $0.7$} & \multicolumn{3}{c}{$>\!75\%$ hidden} & \multicolumn{3}{c}{$>\!90\%$ hidden} \\
\cmidrule(lr){2-4}\cmidrule(lr){5-7}\cmidrule(lr){8-10}\cmidrule(lr){11-13}\cmidrule(lr){14-16}\cmidrule(lr){17-19}
& NN & $k$NN & predicted & NN & $k$NN & predicted & NN & $k$NN & predicted & NN & $k$NN & predicted & NN & $k$NN & predicted & NN & $k$NN & predicted \\
\midrule
MSM$^{\S}$~\cite{hermosilla2025masked} & \textcolor{gray}{53.3} & \textcolor{gray}{53.7} & 50.6 & \textcolor{gray}{41.8} & \textcolor{gray}{42.3} & 39.3 & \textcolor{gray}{45.0} & \textcolor{gray}{45.4} & 43.2 & \textcolor{gray}{32.0} & \textcolor{gray}{32.4} & 29.7 & \textcolor{gray}{34.2} & \textcolor{gray}{34.6} & 39.6 & \textcolor{gray}{25.3} & \textcolor{gray}{26.2} & 33.1 \\
Sonata~\cite{wu2025sonata} & \textcolor{gray}{54.4} & \textcolor{gray}{54.6} & -- & \textcolor{gray}{44.6} & \textcolor{gray}{44.9} & -- & \textcolor{gray}{42.1} & \textcolor{gray}{42.6} & -- & \textcolor{gray}{33.0} & \textcolor{gray}{33.2} & -- & \textcolor{gray}{33.3} & \textcolor{gray}{34.1} & -- & \textcolor{gray}{22.7} & \textcolor{gray}{23.7} & -- \\
Concrt.~\cite{zhang2025concerto} & \textcolor{gray}{59.6} & \textcolor{gray}{59.8} & -- & \textcolor{gray}{48.1} & \textcolor{gray}{48.3} & -- & \textcolor{gray}{48.4} & \textcolor{gray}{48.6} & -- & \textcolor{gray}{36.9} & \textcolor{gray}{37.1} & -- & \textcolor{gray}{38.2} & \textcolor{gray}{39.3} & -- & \textcolor{gray}{27.8} & \textcolor{gray}{29.7} & -- \\
Utonia~\cite{zhang2026utonia} & \textcolor{gray}{60.4} & \textcolor{gray}{60.6} & -- & \textcolor{gray}{49.5} & \textcolor{gray}{49.7} & -- & \textcolor{gray}{51.0} & \textcolor{gray}{50.8} & -- & \textcolor{gray}{37.7} & \textcolor{gray}{37.9} & -- & \textcolor{gray}{41.0} & \textcolor{gray}{41.3} & -- & \textcolor{gray}{29.6} & \textcolor{gray}{29.9} & -- \\
\method{} & \textcolor{gray}{60.2} & \textcolor{gray}{60.3} & \cellcolor[HTML]{def3e4}\bf 64.4 & \textcolor{gray}{49.8} & \textcolor{gray}{49.7} & \cellcolor[HTML]{def3e4}\bf 59.6 & \textcolor{gray}{52.0} & \textcolor{gray}{52.0} & \cellcolor[HTML]{def3e4}\bf 61.7 & \textcolor{gray}{38.2} & \textcolor{gray}{38.4} & \cellcolor[HTML]{def3e4}\bf 51.9 & \textcolor{gray}{39.4} & \textcolor{gray}{39.4} & \cellcolor[HTML]{def3e4}\bf 47.7 & \textcolor{gray}{28.6} & \textcolor{gray}{28.7} & \cellcolor[HTML]{def3e4}\bf 43.5 \\
\midrule
Sonata$^{\dagger}$~\cite{wu2025sonata} & \textcolor{gray}{46.5} & \textcolor{gray}{47.0} & -- & \textcolor{gray}{37.9} & \textcolor{gray}{38.3} & -- & \textcolor{gray}{34.9} & \textcolor{gray}{35.5} & -- & \textcolor{gray}{27.2} & \textcolor{gray}{27.5} & -- & \textcolor{gray}{26.1} & \textcolor{gray}{26.6} & -- & \textcolor{gray}{18.6} & \textcolor{gray}{19.0} & -- \\
\method{}$^{\dagger}$ & \textcolor{gray}{59.5} & \textcolor{gray}{59.6} & \cellcolor[HTML]{def3e4}\bf 62.8 & \textcolor{gray}{48.6} & \textcolor{gray}{48.8} & \cellcolor[HTML]{def3e4}\bf 57.3 & \textcolor{gray}{50.5} & \textcolor{gray}{50.4} & \cellcolor[HTML]{def3e4}\bf 59.7 & \textcolor{gray}{36.4} & \textcolor{gray}{36.6} & \cellcolor[HTML]{def3e4}\bf 49.1 & \textcolor{gray}{37.8} & \textcolor{gray}{37.3} & \cellcolor[HTML]{def3e4}\bf 45.8 & \textcolor{gray}{28.8} & \textcolor{gray}{27.8} & \cellcolor[HTML]{def3e4}\bf 39.3 \\
\bottomrule
\end{tabular}}
\vspace{-5mm}
\end{table*}

\noindent{\bf Comparison with explicit completion.} In \cref{tab:explicit} an off-the-shelf generative scene completer, Seen2Scene~\cite{meng2026seen2scene}, fills the removed regions before each baseline encoder runs, while ours encodes the unfilled scan and queries the predictor at the removed positions. In the single-view columns, the hidden points are scored only on objects more than $75\%$ / $90\%$ hidden from it. Ours performs best in every setting. The generated geometry is often plausible but wrong, and once it enters the scene it also corrupts the features of the visible points around it. Completion is also far more expensive in parameters and time.

\noindent{\bf Cross-domain Generalization: Outdoor Dataset with Limited Viewpoint. } In \cref{tab:transfer} we probe every indoor-pretrained encoder on nuScenes val~\cite{caesar2020nuscenes} and SemanticKITTI val~\cite{behley2019semantickitti}, a domain none of them has seen during pretraining, with the same linear probe. Each LiDAR sweep is sparser and sees the scene from a single viewpoint (ego vehicle). Ours transfers best on both, suggesting that more parameters and more pretraining data do not naturally make features generalize to a different domain.

\noindent{\bf Backbone vs.\ Pretraining.} \cref{tab:cross_backbone} separates the backbone from the pretraining on the same data. With Sonata's pretraining, every backbone degrades sharply as points are removed, including ours, so the shortcut is not specific to an architecture. Replacing it by our pretraining on the same backbone improves every masked column, and the gain grows with the removal ratio. With our pretraining, our backbone leads both Volt models in every column at less than a third of Volt-B's size. The pretraining thus accounts for most of the robustness on every backbone, while PTv3 remains the least robust under either pretraining.

\noindent{\bf Zero-Shot Generalization: Corruptions and Limited Viewpoints.} \cref{tab:corruption} (left) applies three whole-scene corruptions from PointCloud-C~\cite{ren2022benchmarking}, pushed beyond the benchmark's own severity range. Global point dropping is the harshest test: at its severe level every other model loses more than half of its clean performance, whereas ours keeps most of it. \cref{tab:viewpoint} (right) replaces synthetic masks with a common occlusion pattern: a single-viewpoint scene, in which every point outside the view is removed. Here the models distilled from images and larger data come close, and ours is on par with the largest of them at a fraction of their size and without 2D distillation.

\begin{table}[t]
    \centering
        \centering
    \captionsetup{skip=3pt}
    \captionof{table}{\textbf{Emergent occupancy. } AUROC ($\times 100$) of a linear classifier on frozen features, $0.4$\,m cells with $r{=}0.7$; protocol in \cref{sec:eval_protocol}. Whether a hole holds a surface is decodable from our predictor beyond geometric cues, although it was never trained with an empty class or occupancy objective. \\
    \scriptsize{Objects: queries not on walls or floors. Dist.: distance to the nearest visible points. Copied feat.: encoder feature of the nearest visible point.}
    }
    \label{tab:occ}
    \setlength{\tabcolsep}{4pt}
    \footnotesize
    \begin{tabular}{c c c c | c c}
    \toprule
    Pred. feat. & Dist. & xyz & Copied feat. & all \(\uparrow\) & objects \(\uparrow\) \\
    \midrule
    \xmark & \cmark & \xmark & \xmark & 68.2 & 67.8 \\
    \xmark & \cmark & \cmark & \xmark & 68.1 & 67.8 \\
    \xmark & \cmark & \cmark & \cmark & 71.4 & 67.9 \\
    \cellcolor[HTML]{cce3de}\cmark & \cellcolor[HTML]{cce3de}\xmark & \cellcolor[HTML]{cce3de}\xmark & \cellcolor[HTML]{cce3de}\xmark & \cellcolor[HTML]{def3e4}\bf 91.1 & \cellcolor[HTML]{def3e4}\bf 82.2 \\
    \bottomrule
    \vspace{-3 mm}
    \end{tabular}

    \vspace{-4mm}
\end{table}

\begin{figure*}[t]
    \centering
    \captionsetup{skip=3pt}
    \includegraphics[width=1\textwidth]{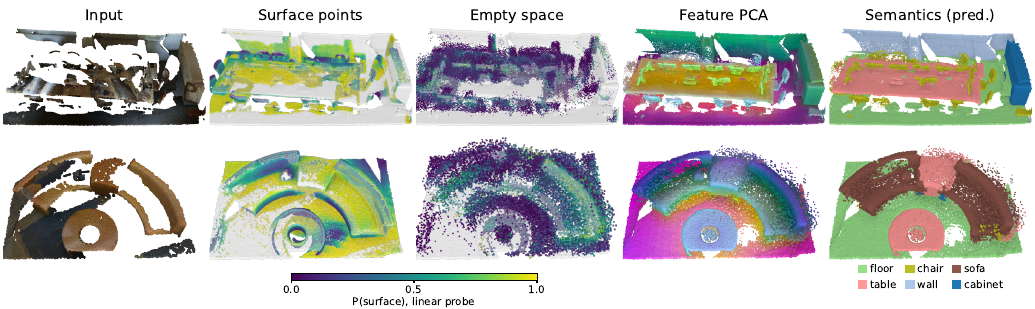}
    \caption{\textbf{Predicting an unobserved region.} Scenes seen from a single viewpoint, so parts of them are hidden. \textit{Surface points} and \textit{empty space:} hidden locations on and off a surface, coloured by how likely our predicted features say a surface is there. \textit{Feature PCA and semantics:} visible and predicted features at the locations recognized as surface (P(surface) > 0.5).}
    \label{fig:occ_field}
    \vspace{-1mm}
\end{figure*}

\begin{table*}[!t]
\captionsetup{skip=3pt}
    \caption{Ablation studies on ScanNet~\cite{dai2017scannet} val. clean: linear-probe mIoU on the complete scan; worst and mean: the largest and the average drop from clean over the six masking settings of \cref{tab:probe}, on its fixed evaluation points; missing in (b): mIoU at removed positions ($0.8$\,m cells, $r{=}0.7$) by $k$NN of encoder features or by the predictor, as in \cref{tab:amodal}. Final selections in \colorbox[HTML]{cce3de}{green}.}
    \label{tab:ablation}
    % Shading: best at one decimal, ties both shaded.
    \begin{minipage}{0.32\textwidth}
    \vspace{-3mm}
    \centering
        \tablestyle{3pt}{1.08}
        \resizebox{\linewidth}{!}{%
        \begin{tabular}{c c | c cc}
        \toprule
        Masked tokens & Pred. & clean \(\uparrow\) & worst \(\uparrow\) & mean \(\uparrow\) \\
        \midrule
        \cmark & \xmark & 69.1 & -43.2 & -21.8 \\
        \xmark & \xmark & 69.5 & -13.0 & -5.8 \\
        \cellcolor[HTML]{cce3de}\xmark & \cellcolor[HTML]{cce3de}\cmark & \cellcolor[HTML]{def3e4}\bf 71.5 & \cellcolor[HTML]{def3e4}\bf -11.5 & \cellcolor[HTML]{def3e4}\bf -4.6 \\
        \bottomrule
        \end{tabular}}
        \subcaption{Masking by removal with predictor.}
        \label{subtab:abl_removal}
        \vspace{-2mm}
    \end{minipage}
    \hfill
    \begin{minipage}{0.32\textwidth}
    \centering
    \scriptsize
        \tablestyle{3pt}{1.08}
        \resizebox{\linewidth}{!}{%
        \begin{tabular}{c c | cc cc}
        \toprule
        \multirow{2}{*}{Pred.} & \multirow{2}{*}{Align.} & \multicolumn{2}{c}{remains} & \multicolumn{2}{c}{missing} \\
        \cmidrule(lr){3-4} \cmidrule(lr){5-6}
        & & clean \(\uparrow\) & mean \(\uparrow\) & $k$NN \(\uparrow\) & pred. \(\uparrow\) \\
        \midrule
        \xmark & \xmark & 69.5 & -5.8 & 34.2 & -- \\
        \cmark & \xmark & 71.2 & -5.0 & 36.2 & 3.6 \\
        \cellcolor[HTML]{cce3de}\cmark & \cellcolor[HTML]{cce3de}\cmark & \cellcolor[HTML]{def3e4}\bf 71.5 & \cellcolor[HTML]{def3e4}\bf -4.6 & \cellcolor[HTML]{def3e4}\bf 36.6 & \cellcolor[HTML]{def3e4}\bf 49.1 \\
        \bottomrule
        \end{tabular}}
        \subcaption{Predictor and alignment.}
        \label{subtab:abl_components}
        \vspace{-2mm}
    \end{minipage}
    \hfill
    \begin{minipage}{0.34\textwidth}
    \centering
        \tablestyle{3pt}{1.08}
        \resizebox{\linewidth}{!}{%
        \begin{tabular}{c c c | c cc}
        \toprule
        P. norm & Out. norm & Ep. & clean \(\uparrow\) & worst \(\uparrow\) & mean \(\uparrow\) \\
        \midrule
        \xmark & \xmark & 100 & 64.5 & -12.2 & -5.5 \\
        \xmark & \cmark & 100 & 64.3 & \cellcolor[HTML]{def3e4}-11.4 & -4.6 \\
        \cmark & \xmark & 100 & 69.7 & -13.2 & -5.6 \\
        \cellcolor[HTML]{cce3de}\cmark & \cellcolor[HTML]{cce3de}\xmark & \cellcolor[HTML]{cce3de}200 & \cellcolor[HTML]{def3e4}71.5 & -11.5 & \cellcolor[HTML]{def3e4}-4.6 \\
        \bottomrule
        \end{tabular}}
        \subcaption{Design choices.}
        \label{subtab:abl_design}
        \vspace{-3mm}
    \end{minipage} \\
    \vspace{2mm}
    \begin{minipage}{0.35\textwidth}
    \centering
        \tablestyle{3pt}{1.08}
        \begin{tabular}{c c c | c cc}
        \toprule
        $\mathcal{L}_{\text{remain}}$ & $\mathcal{L}_{\text{missing}}$ & proto. & clean \(\uparrow\) & worst \(\uparrow\) & mean \(\uparrow\) \\
        \midrule
        \cmark & \xmark & \cmark & 69.5 & -13.0 & -5.8 \\
        \xmark & \cmark & \cmark & 39.2 & -17.3 & -8.9 \\
        \cmark & \cmark & \xmark & \multicolumn{3}{|c}{collapsed} \\
        \cellcolor[HTML]{cce3de}\cmark & \cellcolor[HTML]{cce3de}\cmark & \cellcolor[HTML]{cce3de}\cmark & \cellcolor[HTML]{def3e4}71.5 & \cellcolor[HTML]{def3e4}-11.5 & \cellcolor[HTML]{def3e4}-4.6 \\
        \bottomrule
        \end{tabular}
        \subcaption{Loss terms and prototype assignment.}
        \label{subtab:abl_loss}
    \end{minipage}
    \hfill
    \begin{minipage}{0.345\textwidth}
    \centering
        \tablestyle{3pt}{1.08}
        \begin{tabular}{l c | c cc}
        \toprule
        Sparsest Stage & Params & clean \(\uparrow\) & worst \(\uparrow\) & mean \(\uparrow\) \\
        \midrule
        2 blocks & 22M & 71.1 & -13.6 & -5.2 \\
        \cellcolor[HTML]{cce3de}4 blocks & \cellcolor[HTML]{cce3de}28M & \cellcolor[HTML]{def3e4}71.5 & \cellcolor[HTML]{def3e4}-11.5 & \cellcolor[HTML]{def3e4}-4.6 \\
        \bottomrule
        \end{tabular}
        \subcaption{Encoder design and parameter efficiency.}
        \label{subtab:abl_arch}
    \end{minipage}
    \hfill
    \begin{minipage}{0.27\textwidth}
    \centering
        \tablestyle{3pt}{1.08}
        \begin{tabular}{c | c cc}
        \toprule
        Pred. layer & clean \(\uparrow\) & worst \(\uparrow\) & mean \(\uparrow\) \\
        \midrule
        \cellcolor[HTML]{cce3de}1 & \cellcolor[HTML]{def3e4}\bf 71.5 & \cellcolor[HTML]{def3e4}\bf -11.5 & \cellcolor[HTML]{def3e4}\bf -4.6 \\
        4 & 71.0 & -12.9 & -5.2 \\
        8 & 71.0 & -13.1 & -5.3 \\
        \bottomrule
        \end{tabular}
        \subcaption{Predictor capacity.}
        \label{subtab:abl_pred}
    \end{minipage}
    \vspace{-8mm}
\end{table*}

\subsection{What Is Missing?}
\vspace{-1mm}
\noindent{\bf Semantics at Removed Positions.} \cref{tab:amodal} evaluates semantics at the removed positions. As baselines, the feature at a removed position is copied from the nearest visible point (NN) or averaged over the $k$ nearest ($k$NN). With these interpolations alone, our encoder is on par with the largest models despite its much smaller size, and the best of all once $70\%$ of the scene is removed, the most challenging setting; querying the predictor instead improves further in every setting, and the margin grows with the size of the missing region. MSM~\cite{hermosilla2025masked} is the only open-source baseline with a decoder that can produce features at arbitrary positions. It is designed for training only; we use it at inference to test its predictive ability, querying the removed positions in chunks of $100$, since giving it all of them would already reveal the shape of the missing region. It falls below interpolation of its own features in every setting: the presence shortcut of \cref{sec:motivation}.  See \cref{fig:occ_field} for the visualized examples.

\noindent{\bf Emergent Occupancy.} Our model is never trained to tell whether a hidden location lies on a surface or in empty space: the predictor is only queried on surfaces. Yet after pretraining, this occupancy becomes linearly decodable from the predicted features, well beyond what simple geometric cues provide (\cref{tab:occ}). \cref{fig:occ_field} shows that the hidden surfaces are recognized as such while most of the empty space in front of them is not, and the locations recognized as surface carry the right semantics.

\subsection{Ablation Study}
All variants in \cref{tab:ablation} are pretrained on the ScanNet and ScanNet++ train splits for $102.8$k iterations ($51.4$k for the first three rows of \cref{subtab:abl_design}). Mask tokens at the removed positions keep the clean performance but collapse once points are truly missing (\cref{subtab:abl_removal}), since they still reveal where the geometry is (\cref{sec:additional_exp}). Removal recovers most of the robustness and the predictor improves both further, also on PTv3 and a single-scale Volt (\cref{tab:cross_backbone}). The predictor helps the encoder even without alignment (\cref{subtab:abl_components}); alignment makes its output readable by the linear head and adds robustness. Normalization at the pooling layers is essential: without it, feature magnitudes grow over pretraining and performance drops substantially; normalizing only the output does not help (\cref{subtab:abl_design}). Without $\mathcal{L}_{\text{missing}}$ the model reduces to self-distillation on the visible points and loses both clean performance and robustness; without $\mathcal{L}_{\text{remain}}$, trained only through the predictor, its clean features are far weaker; replacing prototype assignment by feature regression collapses within the first epochs (\cref{subtab:abl_loss}). Halving the sparsest stage changes little (\cref{subtab:abl_arch}); a deeper predictor improves the readout but not the encoder (\cref{subtab:abl_pred}).

\section{Conclusion}

We asked whether self-supervised 3D models, trained to understand masked scenes, stay reliable when points are actually missing. They do not. A mask token at a point's own coordinates leaves its geometry in place, and the pretext task can be solved from that trace. \method removes masked points entirely and supervises the positions they left through a capacity-limited predictor on a hierarchical transformer. The resulting features retain far more of their performance under missing points, corruptions and single-view occlusion, and transfer to sparse outdoor scans, with a fraction of the parameters and two indoor pretraining datasets. Kept after pretraining, the predictor becomes a queryable field of semantic features at locations that were never scanned, with no explicit reconstruction. Our results suggest that closing the shortcut in the pretext task is worth more than the next step in scale.

\clearpage
\noindent{\bf Acknowledgments.}
The Cynergy4MIE project is supported by the Chips Joint Undertaking and its members, including the top-up funding by National Authorities under Grant Agreement No 101140226.

\section*{LLM Usage Statement}
Claude (Anthropic) was used to edit and polish the text, to help implement and run evaluation scripts, and to produce plots. All research ideas, experimental design and conclusions are the authors' own, and the authors verified every number and statement in the paper.

{
    \small
    \bibliographystyle{ieeenat_fullname}
    \bibliography{main}
}
\clearpage
\maketitlesupplementary

\setcounter{section}{0}
\renewcommand{\thesection}{\Alph{section}}
\renewcommand{\theHsection}{supp.\Alph{section}} % separate link anchors from the main-paper sections 1--6

\noindent This supplement backs the main paper with details and further results. \cref{sec:env_data} lists the software environment and the data licenses, and \cref{sec:model_details} the details of the model and its pretraining. \cref{sec:why_hierarchy} puts numbers to the cost argument for a hierarchy, and \cref{sec:eval_protocol} specifies the evaluation protocols. \cref{sec:additional_exp} reports additional results: more datasets, supervised and augmentation baselines, the ablation in full, and controls for the explicit-completion baseline and for querying without oracle positions. \cref{sec:more_vis} shows more visualizations.

\section{Environment and Data Usage}
\label{sec:env_data}

\paragraph{Software Environment and Hardware Resources.}
\begin{itemize}
    \item \textbf{OS:} Ubuntu 24.04.3 LTS (Noble Numbat)
    \item \textbf{Python version:} 3.10.19
    \item \textbf{PyTorch version:} 2.9.1+cu128 (built against CUDA 12.8)
    \item \textbf{Vision library:} torchvision 0.24.1
    \item \textbf{CUDA / cuDNN version:} 12.8 / 9.10.2
    \item \textbf{GPU:} NVIDIA A100 80GB $\times$ 2 for training; NVIDIA A100 80GB $\times$ 1 for evaluation.
\end{itemize}

\paragraph{Data Licenses.} 
In our experiments, we use the publicly available datasets ScanNet~\cite{dai2017scannet} with its ScanNet200 label set~\cite{rozenberszki2022language}, ScanNet++~\cite{yeshwanth2023scannet++}, nuScenes~\cite{caesar2020nuscenes} and SemanticKITTI~\cite{behley2019semantickitti}. 
The ScanNet dataset, including the ScanNet200 annotations, is provided under a custom Terms of Use from the Stanford, Princeton, and TUM organizers, which restricts usage to non-commercial research and educational purposes. 
The ScanNet++ dataset is provided by the Technical University of Munich under the ScanNet++ Terms of Use, which likewise restricts usage to non-commercial research and educational purposes. 
The nuScenes dataset is used under the Creative Commons Attribution-NonCommercial-ShareAlike 4.0 International (CC BY-NC-SA 4.0) license. 
The SemanticKITTI dataset, which builds on the KITTI Vision Benchmark, is distributed under the same CC BY-NC-SA 4.0 license. 
We have strictly adhered to the licensing agreements and terms of use of all datasets.

\section{Model Details}
\label{sec:model_details}
\subsection{Parameter Details.}
\begin{figure}[t]
  \centering
   \includegraphics[width=\linewidth]{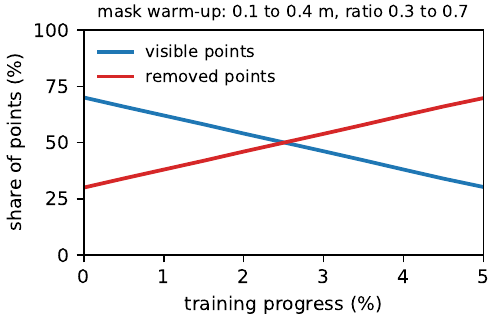}
   \caption{Share of points supervised by each loss over the first $5\%$ of training, the mask warm-up, measured on ScanNet val: visible points are supervised by $\mathcal{L}_{\text{remain}}$, removed points through the predictor by $\mathcal{L}_{\text{missing}}$. Early training is dominated by $\mathcal{L}_{\text{remain}}$; as masks grow, most of the scene is supervised by $\mathcal{L}_{\text{missing}}$.}
   \label{fig:mask_shares}
\end{figure}

\paragraph{Encoder.} The encoder is a hierarchical windowed transformer with four
resolution levels (2, 4, 8 and 16\,cm voxels, stride 2 between levels), channel
widths (32, 192, 384, 512) and depths (0, 6, 6, 4); the 2\,cm level is a linear stem
without attention blocks; LitePT~\cite{yuelitept2026} likewise observes that attention is not needed at every stage, in particular not at the early, high-resolution ones. Every attention block uses 64-dimensional heads
(3, 6 and 8 heads at the three attention levels), MLP ratio 4, QK-normalization,
3D axial rotary position embeddings, and stochastic depth 0.3. The 4\,cm and
8\,cm levels attend within $16^3$-cell windows (0.64\,m and 1.28\,m),
alternating between shifted and unshifted partitions, while the 16\,cm level
attends globally over the scene, which a room-scale scene in ScanNet~\cite{dai2017scannet} affords. Global attention here is a choice, not a requirement: for larger scenes, this level can attend within the same shifted $16^3$-cell windows as the other levels. A LayerNorm follows the stem and every pooling
projection. The point-wise output is the concatenation of all four levels
up-cast to the input resolution (1120 channels). The deployed encoder has 28.1M parameters. The input is grid-sampled to $2$\,cm, keeping one random point per occupied voxel during training; at test time, several such passes cover every point. Pooling projects the $2{\times}2{\times}2$ children of a cell, empty ones filled with zeros, with one linear layer, a stride-$2$ sparse convolution. Up-casting broadcasts each cell's feature to the $2$\,cm voxels it contains, without interpolation, and every point reads the feature of its voxel. During pretraining, a masked point without a visible ancestor cell receives a learned null vector at that level instead; this never occurs at inference.

\paragraph{Masking.} Masks are cubic cells chosen at random. Both the cell size and
the masking ratio are warmed up linearly over the first 5\% of iterations, from
0.1\,m to 0.4\,m and from 0.3 to 0.7, and then held fixed. Cell boundaries are
jittered by 0.01\,m. Masked points are \emph{removed} from the student's input
rather than replaced by mask tokens. \cref{fig:mask_shares} shows the resulting split of supervision: over the warm-up the share of points supervised by $\mathcal{L}_{\text{missing}}$ rises from $30\%$ to $70\%$ (ScanNet val).

\paragraph{Augmentation.} We follow the augmentation of
Sonata~\cite{wu2025sonata}. Each input is augmented 
with random $z$-rotation, small $x/y$ tilts
($\pm2.8^\circ$), scaling in $[0.9, 1.1]$, flips, jitter ($\sigma=5$\,mm) and
elastic distortion.

\paragraph{Self-distillation.} We follow the self-distillation recipe of Sonata~\cite{wu2025sonata}. A teacher, an exponential moving average of the student, encodes unmasked views of the scene; projection heads map student and teacher features onto learned prototypes, and the teacher targets are balanced by Sinkhorn--Knopp. Each scene yields several global and local crops, each with its own augmentation, and only the student's global views are masked. Student and teacher points are paired by their nearest neighbour in the original scene coordinates, so a removed point and its target are the same scan point. $\mathcal{L}_{\text{remain}}$ applies the prototype loss $\ell_{\text{vis}}$ between each masked global view and the teacher's encoding of the same view and of the other global view, and between the local and the global views. $\mathcal{L}_{\text{missing}}$ applies $\ell_{\text{mask}}$ (below) to the removed points written by the predictor and is ramped in by $w(t)$ (\cref{sec:ssl}). Removed points without a visible coarser cell, nearly all of them once the masks reach their final size, receive both terms of $\ell_{\text{mask}}$; the few that inherit a feature from a visible coarser cell receive only the prototype term.

\paragraph{Predictor.} The capacity-limited predictor is a single, narrow cross-attention layer, far smaller than the encoder. Each query is a learned token plus a positional embedding of its target location; the keys and values are the encoder features of the visible points, projected down to the predictor width, and the keys receive the same positional embedding of their own location. On top of it, 3D axial rotary embeddings over $x$, $y$ and $z$ encode the query--key offsets, and queries do not attend to each other. In pretraining, every removed point of a masked view is a query and every visible voxel of the same view a key; at inference the keys are the visible voxels of the scan. The output is projected back to the encoder width through a LayerScale initialised near zero and is supervised by $\ell_{\text{mask}}$ (below).

\paragraph{Feature Alignment.} For a removed position $\mathbf{p}_q$ with predicted feature $\hat{\mathbf{z}}_q$ and teacher feature $\bar{\mathbf{z}}_q$ at the same position, the masked loss of \cref{sec:ssl}, with its per-channel standardization written out, is
\begin{equation}
    \ell_{\text{mask}}\big(\hat{\mathbf{z}}_q, \bar{\mathbf{z}}_q\big) = \ell_{\text{vis}}\big(\hat{\mathbf{z}}_q, \bar{\mathbf{z}}_q\big) + \lambda\, \big\| \tfrac{\hat{\mathbf{z}}_q - \boldsymbol{\mu}}{\boldsymbol{\sigma}} - \tfrac{\bar{\mathbf{z}}_q - \boldsymbol{\mu}}{\boldsymbol{\sigma}} \big\|_{\text{s}\ell_1},
\end{equation}
where $\ell_{\text{vis}}$ is the prototype assignment loss of the visible points, $\boldsymbol{\mu}, \boldsymbol{\sigma}$ are the per-channel mean and standard deviation of the teacher's features in the batch (the latter floored at $0.1\times$ its median), $\|\cdot\|_{\text{s}\ell_1}$ is the smooth-$\ell_1$ distance ($\beta{=}1$) averaged over the $1120$ channels, and $\lambda$ weights the alignment term. The teacher's features are not differentiated. The prototype term only asks that a hidden position fall into the right cluster; the alignment term asks the predictor to return the feature itself, in the teacher's feature space, so that the same semantic head that reads visible points can read a position that was never scanned.

\paragraph{Optimization.} We pretrain on ScanNet~\cite{dai2017scannet} and
ScanNet++~\cite{yeshwanth2023scannet++} (2{,}569 scenes, all splits; no labels
are used) for 400 passes over the data (128k iterations; an epoch in \cref{subtab:abl_design} counts two passes, so this is its $200$-epoch schedule) with a total batch size
of 8 scenes on two NVIDIA A100 (80\,GB) GPUs, in bfloat16 mixed precision.
We use AdamW with a peak learning rate of 0.004 under a OneCycle schedule
(5\% warm-up, cosine annealing, initial and final division factors of 10 and
1000), weight decay increased from 0.04 to 0.1 along a cosine schedule, and
gradient clipping at 3.0.

\section{Why a Hierarchy}
\label{sec:why_hierarchy}
\cref{sec:sfc} argues that a single-scale Transformer must trade resolution against cost. Here we put numbers to that argument, with the Volt~\cite{yilmaz2026volt} as the single-scale reference: it tokenises a scene into $10$\,cm patches of $2$\,cm voxels and applies $12$ blocks of global attention at one width ($384$ for Volt-S, $768$ for Volt-B). Token counts are measured on the $312$ ScanNet~\cite{dai2017scannet} val scenes. FLOPs are computed from them per scene with the standard per-block count, $24Nd^2$ for the linear layers and $4d\sum_w n_w^2$ for attention over windows of $n_w$ tokens ($4dN^2$ when global), and then averaged, one multiply-add counted as two FLOPs. A block thus holds $12d^2$ weights (QKV and output projections, a $4d$ MLP), plus under $1\%$ in biases and norms; the pooling into a level is a linear layer over the $2^3$ child cells, with $8\,d_{\text{in}}d_{\text{out}}$ weights. Parameter counts of models we did not train follow from the same counts, plus, for Volt, its linear tokenizer and voxel decoder ($2.4$M for Volt-S).

\begin{table}[h]
\tablestyle{3pt}{1.05}
\caption{\textbf{Cost of a single-scale encoder} at our deepest width. GFLOPs per ScanNet val scene, mean and largest scene; parameters of Volt~\cite{yilmaz2026volt} variants we did not train and of the flat models are analytic.}
\label{tab:hier_cost}
\resizebox{\linewidth}{!}{
\begin{tabular}{l c c c r r r}
\toprule
Model & tokens & $d$ & blocks & params & GFLOPs & max scene \\
\midrule
\method{} & $4$--$16$\,cm & 192--512 & 16 & 28.1M & 715 & 2,231 \\
Volt-S & $10$\,cm & 384 & 12 & 23.7M & 1,437 & 7,493 \\
Volt-S, $d{=}512$ & $10$\,cm & 512 & 12 & $\approx$41M & 2,037 & 10,322 \\
Volt-B & $10$\,cm & 768 & 12 & 94.0M & 3,518 & 16,752 \\
flat $4$\,cm, global & $4$\,cm & 512 & 16 & $\approx$50M & 68,890 & 395,940 \\
flat $2$\,cm, global & $2$\,cm & 512 & 16 & $\approx$50M & 511,830 & 2.9M \\
\bottomrule
\end{tabular}}
\end{table}

\paragraph{Resolution versus compute.} A single-scale model pays for finer tokens quadratically, since global attention grows with the square of the token count. Moving Volt-S~\cite{yilmaz2026volt} from $10$ to $4$\,cm tokens at our deepest width costs about $100\times$ our compute on a mean scene, and a flat $2$\,cm model is out of reach (\cref{tab:hier_cost}). A hierarchy attends within windows at the fine levels, so their cost grows linearly with the scene.

\paragraph{Width only where it pays.} A flat model gives every token the same width. Raising Volt-S~\cite{yilmaz2026volt} to $d{=}512$ at every block raises its block parameters by $1.8\times$ and its compute by $1.4\times$, to $2.8\times$ ours, because the many fine tokens pay the full width as well. Yet fine tokens describe local geometry and need fewer channels: over $90\%$ of our tokens sit at $2$ and $4$\,cm, with widths $32$ and $192$, and the width of $512$ is spent only at $16$\,cm, where a mean scene has about $2.7$k tokens.

\section{Evaluation Protocol}
\label{sec:eval_protocol}
A result is only as trustworthy as the protocol that produced it~\cite{chang2025seeing, chang2026probing, wang2025uni, chang2024mikasa}, and in 3D vision a sound protocol is particularly hard to design~\cite{chang20253d}: scenes are irregular and unevenly sampled, a few large classes hold most of the points, and any change to the input also changes which points can be scored. This section details the protocols behind our tables.

\paragraph{Fixed Evaluation Points.} A careful choice of the scored points is essential for this evaluation to be meaningful. A naive protocol would score the model on whatever points remain after removal. However, we observe that random removal disproportionately eliminates small and difficult objects, which can vanish entirely even at moderate mask ratios, while large and easy classes such as walls and floors largely survive. Scoring on the remaining points therefore shifts the evaluation toward easier categories and can make the task appear easier as more points are removed. To avoid this bias, we fix the evaluation points: a deterministic $1\%$ of the points in each scene ($358$k points over the $312$ val scenes) is exempt from removal and is the only set scored. The same points are used across all mask sizes and ratios, and for the clean column, so differences in performance reflect only the loss of surrounding context rather than a change in which points are evaluated. The evaluation points inside removed cells leave only a faint trace of the missing surface: $3.9\%$ of the $4$\,cm voxels of removed cells stay occupied.

\paragraph{Mostly Hidden Objects.} Here removed cells are fully empty and the evaluation points are removed with their cells. For a removal ratio $r$, $\lceil 1/(1-r)\rceil$ masks are drawn so that every cell survives in at least one of them, and each of the fixed evaluation points is scored once, in the first mask where its own cell survives. The hidden fraction of its object is the share of the object's points removed in that mask, for instances with at least $100$ points, walls and floors excluded. Each bin is also scored on the same points in the complete scan, and settings with fewer than $100$ objects in a bin are omitted.

\paragraph{Single Viewpoint.} The camera is synthetic and panoramic, $1.6$\,m above the floor. It is placed inside the room and biased toward a corner: candidates offset from the scene centre toward a random corner by $25$--$40\%$ of the extent are accepted if no geometry lies within $0.3$\,m at eye height and the floor lies within $0.4$\,m below. Visibility is computed on the scan points by spherical-flip hidden-point removal, and hidden points are removed from the input. The hidden fraction of an object is the share of its points that are hidden, for instances with at least $100$ points of which at least $20$ are visible, walls and floors excluded. The same camera is used in \cref{tab:viewpoint,tab:amodal-vp,tab:explicit}. \cref{tab:viewpoint} scores the visible points of the selected objects and \cref{tab:amodal-vp} their hidden points; \cref{tab:explicit} scores all visible points, walls and floors included, together with the hidden points of the selected objects.

\paragraph{Missing-Point Queries.} \cref{tab:amodal} scores $4{,}000$ removed labelled points per val scene, sampled uniformly without replacement with a fixed seed per scene, or all of them if fewer; under a single viewpoint, they are sampled from the hidden points of objects more than $75\%$ hidden, and the $90\%$ column is the subset of those. $k$NN is the plain mean of the $k{=}8$ nearest visible features, NN its $k{=}1$ case. The decoder of MSM~\cite{hermosilla2025masked} is queried on the same points, $100$ at a time. Its decoder processes all queries jointly, so queried together they would reveal the outline of the removed geometry; small groups limit this leak, and the remaining leak only favours MSM.

% Shading as in tab:probe, computed per dataset block: best (dark, bold) / second (light) among the first five rows, best only within the dagger pair, clean unshaded.
\begin{table*}[t]
    \centering
    \captionsetup{skip=3pt}
    \caption{\textbf{Robustness to missing points on ScanNet200 and ScanNet++.} As \cref{tab:probe}, with a linear head trained on the same frozen backbones with each model's own probe recipe: for the $200$ classes of ScanNet200~\cite{rozenberszki2022language}, and for the top-$100$ semantic classes of ScanNet++~\cite{yeshwanth2023scannet++}, trained on its $856$ train scenes and scored on its $50$ val scenes. The object blocks add $0.4$@$50$; $0.2$@$50$ leaves fewer than $100$ such objects on both datasets and is omitted. On ScanNet++, the rows without $^{\dagger}$ include its val scenes in pretraining; MSM is not pretrained on ScanNet++ and is not evaluated there. $^{\S}$, $^{\dagger}$: as in \cref{tab:probe}.}
    \label{tab:probe200}
    \label{tab:probe_snpp}
    \footnotesize
    \setlength{\tabcolsep}{2.5pt}
    \resizebox{\linewidth}{!}{%
    \begin{tabular}{l | c cccccc | ccccc || c cccccc | ccccc}
    \toprule
    \multirow{3}{*}{Model} & \multicolumn{12}{c||}{ScanNet200} & \multicolumn{12}{c}{ScanNet++} \\
    \cmidrule(lr){2-13}\cmidrule(lr){14-25}
    & \multicolumn{7}{c|}{Random mask (fixed $1\%$ scored)} & \multicolumn{5}{c||}{Objects $>\!90\%$ hidden} & \multicolumn{7}{c|}{Random mask (fixed $1\%$ scored)} & \multicolumn{5}{c}{Objects $>\!90\%$ hidden} \\
    \cmidrule(lr){2-8}\cmidrule(lr){9-13}\cmidrule(lr){14-20}\cmidrule(lr){21-25}
    & \textcolor{gray}{clean} & {\scriptsize 0.2@50} & {\scriptsize 0.4@50} & {\scriptsize 0.2@70} & {\scriptsize 0.4@70} & {\scriptsize 0.2@90} & {\scriptsize 0.4@90} & {\scriptsize 0.4@50} & {\scriptsize 0.2@70} & {\scriptsize 0.4@70} & {\scriptsize 0.2@90} & {\scriptsize 0.4@90} & \textcolor{gray}{clean} & {\scriptsize 0.2@50} & {\scriptsize 0.4@50} & {\scriptsize 0.2@70} & {\scriptsize 0.4@70} & {\scriptsize 0.2@90} & {\scriptsize 0.4@90} & {\scriptsize 0.4@50} & {\scriptsize 0.2@70} & {\scriptsize 0.4@70} & {\scriptsize 0.2@90} & {\scriptsize 0.4@90} \\
    \midrule
    MSM$^{\S}$~\cite{hermosilla2025masked} & \textcolor{gray}{25.0} & 22.0 & 21.1 & 19.0 & 17.7 & \cellcolor[HTML]{eef9f0}12.8 & \cellcolor[HTML]{eef9f0}12.2 & 14.4 & 14.0 & 16.3 & 12.4 & 10.9 & -- & -- & -- & -- & -- & -- & -- & -- & -- & -- & -- & -- \\
    Sonata~\cite{wu2025sonata} & \textcolor{gray}{28.3} & 24.9 & 21.8 & 19.1 & 14.0 & 6.1 & 5.0 & 11.9 & 10.6 & 13.0 & 5.2 & 7.4 & \textcolor{gray}{38.1} & 32.3 & 30.4 & 26.8 & 24.7 & 20.9 & 19.1 & 20.1 & 16.4 & 18.9 & 13.4 & 16.1 \\
    Concrt.~\cite{zhang2025concerto} & \textcolor{gray}{35.6} & \cellcolor[HTML]{eef9f0}29.5 & \cellcolor[HTML]{eef9f0}26.2 & 20.7 & 17.2 & 8.2 & 7.0 & 16.6 & 15.2 & 16.0 & 10.0 & 12.5 & \textcolor{gray}{43.0} & \cellcolor[HTML]{eef9f0}34.3 & 31.4 & 28.9 & 26.7 & 22.5 & 20.6 & 19.9 & 16.9 & 21.5 & 15.9 & 18.2 \\
    Utonia~\cite{zhang2026utonia} & \textcolor{gray}{35.0} & \cellcolor[HTML]{def3e4}\bf 29.8 & \cellcolor[HTML]{def3e4}\bf 28.4 & \cellcolor[HTML]{eef9f0}20.9 & \cellcolor[HTML]{eef9f0}21.2 & 11.1 & 10.1 & \cellcolor[HTML]{def3e4}\bf 19.9 & \cellcolor[HTML]{eef9f0}16.1 & \cellcolor[HTML]{def3e4}\bf 22.7 & \cellcolor[HTML]{eef9f0}14.5 & \cellcolor[HTML]{def3e4}\bf 15.1 & \textcolor{gray}{43.2} & 34.1 & \cellcolor[HTML]{eef9f0}31.9 & \cellcolor[HTML]{eef9f0}29.6 & \cellcolor[HTML]{eef9f0}27.4 & \cellcolor[HTML]{eef9f0}24.7 & \cellcolor[HTML]{eef9f0}22.8 & \cellcolor[HTML]{eef9f0}22.5 & \cellcolor[HTML]{eef9f0}19.3 & \cellcolor[HTML]{def3e4}\bf 25.1 & \cellcolor[HTML]{eef9f0}20.3 & \cellcolor[HTML]{eef9f0}21.3 \\
    \method{} & \textcolor{gray}{28.8} & 27.5 & 26.2 & \cellcolor[HTML]{def3e4}\bf 25.8 & \cellcolor[HTML]{def3e4}\bf 23.3 & \cellcolor[HTML]{def3e4}\bf 19.9 & \cellcolor[HTML]{def3e4}\bf 18.0 & \cellcolor[HTML]{eef9f0}19.1 & \cellcolor[HTML]{def3e4}\bf 18.2 & \cellcolor[HTML]{eef9f0}20.6 & \cellcolor[HTML]{def3e4}\bf 17.6 & \cellcolor[HTML]{eef9f0}15.0 & \textcolor{gray}{38.4} & \cellcolor[HTML]{def3e4}\bf 36.1 & \cellcolor[HTML]{def3e4}\bf 34.4 & \cellcolor[HTML]{def3e4}\bf 32.4 & \cellcolor[HTML]{def3e4}\bf 30.6 & \cellcolor[HTML]{def3e4}\bf 27.3 & \cellcolor[HTML]{def3e4}\bf 25.6 & \cellcolor[HTML]{def3e4}\bf 28.2 & \cellcolor[HTML]{def3e4}\bf 19.4 & \cellcolor[HTML]{eef9f0}24.7 & \cellcolor[HTML]{def3e4}\bf 23.3 & \cellcolor[HTML]{def3e4}\bf 22.8 \\
    \midrule
    Sonata$^{\dagger}$~\cite{wu2025sonata} & \textcolor{gray}{23.8} & 19.6 & 17.1 & 12.9 & 8.4 & 2.8 & 2.1 & 10.7 & 6.9 & 6.6 & 2.7 & 4.1 & \textcolor{gray}{34.4} & 29.3 & 27.0 & 24.6 & 21.7 & 17.5 & 14.7 & 15.7 & 10.7 & 14.9 & 7.4 & 10.2 \\
    \method{}$^{\dagger}$ & \textcolor{gray}{28.3} & \cellcolor[HTML]{def3e4}\bf 27.0 & \cellcolor[HTML]{def3e4}\bf 25.7 & \cellcolor[HTML]{def3e4}\bf 25.0 & \cellcolor[HTML]{def3e4}\bf 22.5 & \cellcolor[HTML]{def3e4}\bf 19.3 & \cellcolor[HTML]{def3e4}\bf 17.6 & \cellcolor[HTML]{def3e4}\bf 14.2 & \cellcolor[HTML]{def3e4}\bf 14.0 & \cellcolor[HTML]{def3e4}\bf 19.8 & \cellcolor[HTML]{def3e4}\bf 15.4 & \cellcolor[HTML]{def3e4}\bf 12.3 & \textcolor{gray}{37.8} & \cellcolor[HTML]{def3e4}\bf 35.7 & \cellcolor[HTML]{def3e4}\bf 34.8 & \cellcolor[HTML]{def3e4}\bf 32.8 & \cellcolor[HTML]{def3e4}\bf 30.8 & \cellcolor[HTML]{def3e4}\bf 27.4 & \cellcolor[HTML]{def3e4}\bf 25.5 & \cellcolor[HTML]{def3e4}\bf 24.2 & \cellcolor[HTML]{def3e4}\bf 20.7 & \cellcolor[HTML]{def3e4}\bf 23.9 & \cellcolor[HTML]{def3e4}\bf 22.8 & \cellcolor[HTML]{def3e4}\bf 20.3 \\
    \bottomrule
    \end{tabular}}
\end{table*}

\begin{table*}[t]
\centering
\captionsetup{skip=3pt}
\caption{\textbf{Per-class IoU of the outdoor transfer.} The linear heads of \cref{tab:transfer} on nuScenes val (top) and SemanticKITTI sequence 08 (bottom); mIoU as in \cref{tab:transfer}.}
\label{tab:perclass_transfer}
\footnotesize
\setlength{\tabcolsep}{2.6pt}
\begin{tabular}{l cccccccccccccccc | c}
\toprule
nuScenes & \rotatebox{90}{barrier} & \rotatebox{90}{bicycle} & \rotatebox{90}{bus} & \rotatebox{90}{car} & \rotatebox{90}{constr. vehicle} & \rotatebox{90}{motorcycle} & \rotatebox{90}{pedestrian} & \rotatebox{90}{traffic cone} & \rotatebox{90}{trailer} & \rotatebox{90}{truck} & \rotatebox{90}{driveable} & \rotatebox{90}{other flat} & \rotatebox{90}{sidewalk} & \rotatebox{90}{terrain} & \rotatebox{90}{man-made} & \rotatebox{90}{vegetation} & mIoU \\
\midrule
MSM~\cite{hermosilla2025masked} & 45.8 & 0.0 & 12.8 & 60.7 & 0.0 & 17.4 & 23.3 & 14.4 & 1.6 & 26.6 & 78.5 & 11.1 & 19.5 & 40.1 & 67.3 & 65.3 & 30.3 \\
Sonata~\cite{wu2025sonata} & 50.5 & 5.4 & \cellcolor[HTML]{def3e4}\bf 47.2 & 69.3 & 7.2 & 31.9 & \cellcolor[HTML]{def3e4}\bf 35.8 & \cellcolor[HTML]{def3e4}\bf 24.3 & 13.1 & 40.5 & 85.1 & 34.3 & 34.4 & \cellcolor[HTML]{def3e4}\bf 47.4 & 73.4 & 69.7 & 41.8 \\
Concrt.~\cite{zhang2025concerto} & 51.8 & 5.4 & 38.3 & 69.4 & 9.1 & \cellcolor[HTML]{def3e4}\bf 32.9 & 33.2 & 23.5 & 11.6 & 40.6 & 84.5 & 30.1 & 34.7 & 44.1 & \cellcolor[HTML]{def3e4}\bf 73.7 & \cellcolor[HTML]{def3e4}\bf 70.6 & 40.9 \\
\method{} & \cellcolor[HTML]{def3e4}\bf 55.8 & \cellcolor[HTML]{def3e4}\bf 7.4 & 46.2 & \cellcolor[HTML]{def3e4}\bf 70.2 & \cellcolor[HTML]{def3e4}\bf 13.1 & \cellcolor[HTML]{def3e4}\bf 32.9 & 31.5 & 22.0 & \cellcolor[HTML]{def3e4}\bf 15.3 & \cellcolor[HTML]{def3e4}\bf 47.1 & \cellcolor[HTML]{def3e4}\bf 87.1 & \cellcolor[HTML]{def3e4}\bf 41.1 & \cellcolor[HTML]{def3e4}\bf 41.2 & 43.8 & 69.2 & 66.1 & \cellcolor[HTML]{def3e4}\bf 43.1 \\
\bottomrule
\end{tabular}
\vspace{4pt}

\begin{tabular}{l ccccccccccccccccccc | c}
\toprule
SemanticKITTI & \rotatebox{90}{car} & \rotatebox{90}{bicycle} & \rotatebox{90}{motorcycle} & \rotatebox{90}{truck} & \rotatebox{90}{other-vehicle} & \rotatebox{90}{person} & \rotatebox{90}{bicyclist} & \rotatebox{90}{motorcyclist} & \rotatebox{90}{road} & \rotatebox{90}{parking} & \rotatebox{90}{sidewalk} & \rotatebox{90}{other-ground} & \rotatebox{90}{building} & \rotatebox{90}{fence} & \rotatebox{90}{vegetation} & \rotatebox{90}{trunk} & \rotatebox{90}{terrain} & \rotatebox{90}{pole} & \rotatebox{90}{traffic-sign} & mIoU \\
\midrule
MSM~\cite{hermosilla2025masked} & 77.1 & 0.0 & 8.7 & 16.2 & 6.9 & 0.0 & 0.0 & 0.0 & 62.8 & 8.5 & 43.7 & 0.0 & 74.6 & 25.6 & 80.3 & 46.2 & 53.9 & 43.4 & 16.5 & 29.7 \\
Sonata~\cite{wu2025sonata} & 84.1 & 9.1 & \cellcolor[HTML]{def3e4}\bf 30.3 & 30.5 & 20.3 & \cellcolor[HTML]{def3e4}\bf 40.5 & \cellcolor[HTML]{def3e4}\bf 51.9 & \cellcolor[HTML]{def3e4}\bf 0.3 & 73.7 & \cellcolor[HTML]{def3e4}\bf 15.0 & 53.4 & 0.2 & 82.1 & 37.8 & 83.6 & \cellcolor[HTML]{def3e4}\bf 59.7 & 61.3 & \cellcolor[HTML]{def3e4}\bf 56.4 & 35.8 & 43.5 \\
Concrt.~\cite{zhang2025concerto} & 84.8 & 9.3 & 27.5 & \cellcolor[HTML]{def3e4}\bf 33.5 & 24.3 & 32.5 & 41.4 & \cellcolor[HTML]{def3e4}\bf 0.3 & 71.2 & 13.2 & 51.8 & 0.3 & 81.9 & 35.4 & 83.5 & 58.5 & 56.5 & 56.1 & \cellcolor[HTML]{def3e4}\bf 36.5 & 42.0 \\
\method{} & \cellcolor[HTML]{def3e4}\bf 86.9 & \cellcolor[HTML]{def3e4}\bf 10.0 & 29.5 & 31.6 & \cellcolor[HTML]{def3e4}\bf 30.2 & 39.3 & 40.6 & 0.1 & \cellcolor[HTML]{def3e4}\bf 76.6 & 14.0 & \cellcolor[HTML]{def3e4}\bf 56.6 & \cellcolor[HTML]{def3e4}\bf 5.1 & \cellcolor[HTML]{def3e4}\bf 83.9 & \cellcolor[HTML]{def3e4}\bf 46.0 & \cellcolor[HTML]{def3e4}\bf 84.0 & 54.9 & \cellcolor[HTML]{def3e4}\bf 64.1 & 51.6 & 30.8 & \cellcolor[HTML]{def3e4}\bf 44.0 \\
\bottomrule
\end{tabular}
\end{table*}

\paragraph{Linear Probe.} Every backbone is frozen, and every head is trained on complete ScanNet~\cite{dai2017scannet} train scans, without masking. Ours reads the concatenation of its features across levels, standardized by a parameter-free batch normalization, with one linear layer, trained with the linear-probe recipe of Sonata~\cite{wu2025sonata}. The baselines use their official linear-probe configurations, with the released probe weights where available; MSM~\cite{hermosilla2025masked}, which has none, is evaluated with its own head, a linear layer on its standardized decoder output ($^{\S}$ in \cref{tab:probe}). No test-time augmentation is used anywhere. \cref{tab:corruption} scores all labelled points of each val scene, \cref{tab:probe} only its fixed evaluation points, so their clean numbers differ slightly.

\paragraph{Outdoor Transfer.} In \cref{tab:transfer}, every encoder stays frozen and a linear head is trained on the train split of each outdoor dataset, with one short training schedule shared by all models. nuScenes~\cite{caesar2020nuscenes} is scored over its $16$ classes on its val split, and SemanticKITTI~\cite{behley2019semantickitti} over its $19$ classes on the $4{,}071$ scans of sequence 08, at full resolution. LiDAR sweeps carry no normals, so they are estimated from the points of each sweep, for every model. Utonia~\cite{zhang2026utonia} is left out because its pretraining includes both outdoor datasets.

\paragraph{Occupancy Probe.} \cref{tab:occ} uses one mask, $0.4$\,m cells with $r{=}0.7$. Surface queries are removed points. Empty-space queries are removed points pushed a short distance outward along their scan normal, into the free side of the surface, and kept only if they lie inside the scene and away from every point of the complete scan; they form a band in front of the hidden surfaces rather than filling the scene uniformly. Both classes have equally many queries. Each query carries the class of its source surface point, and the objects column keeps the queries whose source is neither wall nor floor. The probe is a logistic regression on standardized inputs, fit on a subset of ScanNet train scenes and evaluated on the val scenes. Inputs: Dist., the distances from the query to its nearest visible points, computed from the visible points alone, since the removed surface and its normals are unavailable to every model; xyz, the query position in the scene, plus its height above the floor; Copied feat., the encoder feature of the nearest visible point; Pred. feat., the predictor output at the query. For our predictor, AUROC is higher the farther the query lies from the surface.

\section{Additional Results}
\label{sec:additional_exp}

\begin{table}[t]
    \centering
    \captionsetup{skip=3pt}
    \caption{Comparison with supervised methods (ScanNet val mIoU), evaluated with the protocol and fixed evaluation points of \cref{tab:probe}. sup.: trained fully supervised. All models are trained on train splits only.}
    \label{tab:supervised}
    \setlength{\tabcolsep}{2.5pt}
    \small
    \resizebox{\linewidth}{!}{%
    \begin{tabular}{l cc ccc ccc ccc}
    \toprule
    \multirow{2}{*}{Model} & \multirow{2}{*}{sup.} & \multirow{2}{*}{clean} & \multicolumn{3}{c}{$r=0.5$} & \multicolumn{3}{c}{$r=0.7$} & \multicolumn{3}{c}{$r=0.9$} \\
    \cmidrule(lr){4-6}\cmidrule(lr){7-9}\cmidrule(lr){10-12}
    & & & $b{=}0.2$ & $0.4$ & $0.8$ & $0.2$ & $0.4$ & $0.8$ & $0.2$ & $0.4$ & $0.8$ \\
    \midrule
    PTv3~\cite{wu2024point} & \cmark & \textcolor{gray}{75.8} & 66.4 & 57.0 & 48.7 & 53.4 & 41.6 & 34.0 & 23.7 & 17.8 & 17.1 \\
    Volt-S~\cite{yilmaz2026volt} & \cmark & \textcolor{gray}{75.7} & 68.9 & 62.5 & 52.1 & 60.9 & 49.3 & 38.7 & 36.5 & 28.2 & 23.9 \\
    \method{} & \xmark & \textcolor{gray}{71.5} & \cellcolor[HTML]{def3e4}\textbf{71.2} & \cellcolor[HTML]{def3e4}\textbf{69.6} & \cellcolor[HTML]{def3e4}\textbf{67.8} & \cellcolor[HTML]{def3e4}\textbf{69.7} & \cellcolor[HTML]{def3e4}\textbf{67.1} & \cellcolor[HTML]{def3e4}\textbf{63.8} & \cellcolor[HTML]{def3e4}\textbf{64.2} & \cellcolor[HTML]{def3e4}\textbf{60.0} & \cellcolor[HTML]{def3e4}\textbf{56.7} \\
    \bottomrule
    \end{tabular}}
\end{table}

\begin{figure}[!ht]
    \centering
    \captionsetup{skip=5pt}
    \includegraphics[width=1\linewidth]{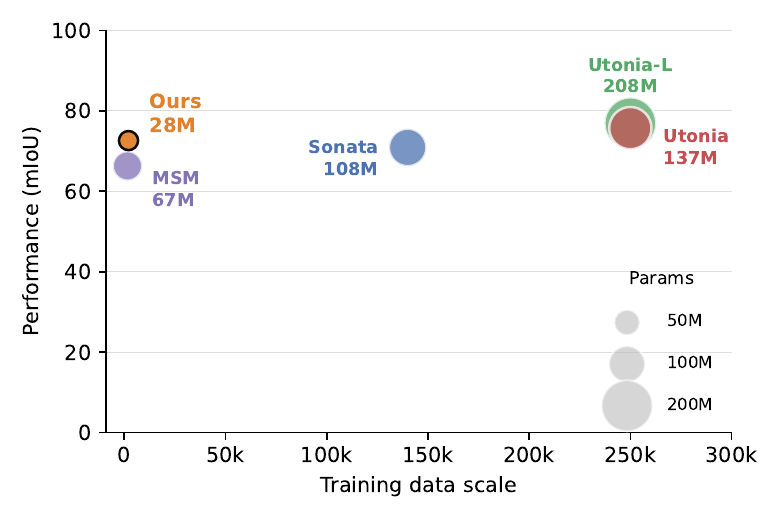}
    \caption{\textbf{Performance versus scale.} Clean ScanNet val mIoU against the number of pretraining scenes; marker area is the parameter count. \method, pretrained on far less data with a fraction of the parameters, stays competitive with models trained on far more.}
    \label{fig:scale_chart}
\end{figure}

\paragraph{Performance versus Scale.} \cref{fig:scale_chart} places MSM~\cite{hermosilla2025masked}, Sonata~\cite{wu2025sonata}, Utonia~\cite{zhang2026utonia}, its larger variant Utonia-L and ours by clean ScanNet~\cite{dai2017scannet} performance against the number of pretraining scenes, with the parameter count as marker area. Far more data and several times the parameters buy at most a few points, and \method, pretrained on far less data with a fraction of the parameters, stays competitive.

\begin{table*}[t]
\centering
\caption{\textbf{Per-class robustness.} Relative change in IoU (\%) from each model's own clean IoU, $100\,(\text{IoU}_{\text{masked}}-\text{IoU}_{\text{clean}})/\text{IoU}_{\text{clean}}$, averaged over the nine cells $b\in\{0.2,0.4,0.8\}$\,m $\times$ $r\in\{0.5,0.7,0.9\}$ of \cref{tab:probe}, on its fixed evaluation points, for all $20$ ScanNet classes; mean (20): the same relative change of the mIoU. Darker red: larger drop. $^{\S}$: as in \cref{tab:probe}.}
\label{tab:perclass}
\setlength{\tabcolsep}{3pt}
\small
\resizebox{\textwidth}{!}{%
\begin{tabular}{l cccccccccccccccccccc | c}
\toprule
Model & \rotatebox{90}{wall} & \rotatebox{90}{floor} & \rotatebox{90}{cabinet} & \rotatebox{90}{bed} & \rotatebox{90}{chair} & \rotatebox{90}{sofa} & \rotatebox{90}{table} & \rotatebox{90}{door} & \rotatebox{90}{window} & \rotatebox{90}{bookshelf} & \rotatebox{90}{picture} & \rotatebox{90}{counter} & \rotatebox{90}{desk} & \rotatebox{90}{curtain} & \rotatebox{90}{refrigerator} & \rotatebox{90}{shower curt.} & \rotatebox{90}{toilet} & \rotatebox{90}{sink} & \rotatebox{90}{bathtub} & \rotatebox{90}{other furn.} & \rotatebox{90}{mean (20)} \\
\midrule
MSM$^{\S}$~\cite{hermosilla2025masked} & \cellcolor{Salmon!29}-9.3 & \cellcolor{Salmon!20}-4.6 & \cellcolor{Salmon!41}-18.6 & \cellcolor{Salmon!40}-17.4 & \cellcolor{Salmon!30}-9.9 & \cellcolor{Salmon!44}-21.3 & \cellcolor{Salmon!43}-20.7 & \cellcolor{Salmon!43}-20.4 & \cellcolor{Salmon!41}-18.5 & \cellcolor{Salmon!24}-6.6 & \cellcolor{Salmon!60}-54.7 & \cellcolor{Salmon!60}-46.6 & \cellcolor{Salmon!36}-14.7 & \cellcolor{Salmon!46}-23.2 & \cellcolor{Salmon!43}-20.8 & \cellcolor{Salmon!33}-12.0 & \cellcolor{Salmon!36}-14.6 & \cellcolor{Salmon!42}-20.0 & \cellcolor{Salmon!46}-23.0 & \cellcolor{Salmon!41}-18.8 & \cellcolor{Salmon!40}-18.1 \\
Sonata~\cite{wu2025sonata} & \cellcolor{Salmon!37}-15.2 & \cellcolor{Salmon!21}-4.8 & \cellcolor{Salmon!55}-33.6 & \cellcolor{Salmon!52}-30.3 & \cellcolor{Salmon!47}-24.0 & \cellcolor{Salmon!60}-39.6 & \cellcolor{Salmon!52}-29.5 & \cellcolor{Salmon!60}-51.0 & \cellcolor{Salmon!55}-33.6 & \cellcolor{Salmon!60}-44.9 & \cellcolor{Salmon!60}-63.3 & \cellcolor{Salmon!60}-39.8 & \cellcolor{Salmon!60}-48.0 & \cellcolor{Salmon!48}-25.7 & \cellcolor{Salmon!54}-32.2 & \cellcolor{Salmon!55}-34.1 & \cellcolor{Salmon!50}-27.9 & \cellcolor{Salmon!52}-30.5 & \cellcolor{Salmon!60}-51.3 & \cellcolor{Salmon!60}-41.3 & \cellcolor{Salmon!55}-33.3 \\
Concerto~\cite{zhang2025concerto} & \cellcolor{Salmon!40}-17.6 & \cellcolor{Salmon!26}-7.4 & \cellcolor{Salmon!54}-32.0 & \cellcolor{Salmon!52}-30.5 & \cellcolor{Salmon!51}-29.0 & \cellcolor{Salmon!60}-50.4 & \cellcolor{Salmon!53}-30.8 & \cellcolor{Salmon!60}-42.8 & \cellcolor{Salmon!58}-37.0 & \cellcolor{Salmon!45}-22.1 & \cellcolor{Salmon!60}-56.3 & \cellcolor{Salmon!57}-36.3 & \cellcolor{Salmon!54}-32.9 & \cellcolor{Salmon!48}-25.9 & \cellcolor{Salmon!60}-43.3 & \cellcolor{Salmon!55}-33.2 & \cellcolor{Salmon!57}-35.9 & \cellcolor{Salmon!60}-40.1 & \cellcolor{Salmon!60}-40.1 & \cellcolor{Salmon!60}-49.0 & \cellcolor{Salmon!55}-33.5 \\
Utonia~\cite{zhang2026utonia} & \cellcolor{Salmon!33}-11.7 & \cellcolor{Salmon!24}-6.4 & \cellcolor{Salmon!45}-22.3 & \cellcolor{Salmon!42}-19.8 & \cellcolor{Salmon!42}-20.0 & \cellcolor{Salmon!44}-21.8 & \cellcolor{Salmon!48}-25.9 & \cellcolor{Salmon!56}-35.1 & \cellcolor{Salmon!48}-25.1 & \cellcolor{Salmon!40}-17.4 & \cellcolor{Salmon!60}-65.3 & \cellcolor{Salmon!40}-17.4 & \cellcolor{Salmon!54}-32.8 & \cellcolor{Salmon!55}-33.4 & \cellcolor{Salmon!52}-30.5 & \cellcolor{Salmon!42}-19.6 & \cellcolor{Salmon!45}-22.2 & \cellcolor{Salmon!48}-26.1 & \cellcolor{Salmon!46}-23.3 & \cellcolor{Salmon!57}-36.1 & \cellcolor{Salmon!47}-24.2 \\
\method{} & \cellcolor{Salmon!17}-3.3 & \cellcolor{Salmon!10}-1.1 & \cellcolor{Salmon!20}-4.4 & \cellcolor{Salmon!24}-6.5 & \cellcolor{Salmon!19}-4.2 & \cellcolor{Salmon!22}-5.4 & \cellcolor{Salmon!24}-6.6 & \cellcolor{Salmon!26}-7.3 & \cellcolor{Salmon!27}-8.0 & \cellcolor{Salmon!15}-2.6 & \cellcolor{Salmon!60}-46.2 & \cellcolor{Salmon!30}-10.2 & \cellcolor{Salmon!30}-9.9 & \cellcolor{Salmon!29}-9.2 & \cellcolor{Salmon!24}-6.1 & \cellcolor{Salmon!28}-8.8 & \cellcolor{Salmon!31}-10.6 & \cellcolor{Salmon!31}-10.4 & \cellcolor{Salmon!34}-13.0 & \cellcolor{Salmon!29}-9.2 & \cellcolor{Salmon!27}-8.0 \\
\bottomrule
\end{tabular}}
% tab:perclass shading: Salmon!min(60, round(9.5 sqrt(-x))).
\end{table*}

\paragraph{What Kind of Objects Suffer the Most?}
\label{sec:supp_perclass}
\cref{tab:perclass} lists the per-class change in IoU when points are removed.
\emph{Picture}, thin and flat against the wall, suffers the most: it drops by $54.7\%$ to
$65.3\%$ for every baseline and is the worst class of every model. Floor changes least for every model, and wall is among its three most stable classes. Ours loses less than $14\%$ on every one of the $20$ classes except picture ($-46.2\%$).

\begin{table*}[t]
    \centering
    \caption{\textbf{Block-removal augmentation is interpolative.} ScanNet val mIoU of the official PTv3, PTv3 retrained with random block removal as augmentation (block sizes $b \le 0.4$\,m, up to half of the blocks removed; the two columns inside that range are marked $^{*}$), and \method{}. All models are trained on train splits only. Block columns use the protocol and fixed evaluation points of \cref{tab:probe}; mean and worst are the change from clean over the twelve block cells. Single view follows \cref{tab:viewpoint} (full scene $\to$ one viewpoint, on the same objects) and drop\_global follows \cref{tab:corruption} (clean on the full val set $\to$ $99\%$ of points removed). The two PTv3 models were evaluated in one run on identical inputs.}
    \label{tab:aug}
    \setlength{\tabcolsep}{3pt}
    \small
    \resizebox{\textwidth}{!}{%
    \begin{tabular}{l c cccc cccc cccc cc cc c}
    \toprule
    \multirow{2}{*}{Model} & \multirow{2}{*}{clean} & \multicolumn{4}{c}{$r=0.5$} & \multicolumn{4}{c}{$r=0.7$} & \multicolumn{4}{c}{$r=0.9$} & \multicolumn{2}{c}{$\Delta$ vs.\ clean} & \multicolumn{2}{c}{single view (full $\to$ 1 view)} & drop\_global \\
    \cmidrule(lr){3-6}\cmidrule(lr){7-10}\cmidrule(lr){11-14}\cmidrule(lr){15-16}\cmidrule(lr){17-18}\cmidrule(lr){19-19}
    & & $b{=}0.2^{*}$ & $0.4^{*}$ & $0.8$ & $1.6$ & $0.2$ & $0.4$ & $0.8$ & $1.6$ & $0.2$ & $0.4$ & $0.8$ & $1.6$ & mean & worst & $>\!75\%$ hidden & $>\!90\%$ hidden & $99\%$ \\
    \midrule
    PTv3~\cite{wu2024point} & \textcolor{gray}{75.8} & 66.4 & 57.0 & 48.7 & 42.4 & 53.4 & 41.6 & 34.0 & 29.0 & 23.7 & 17.8 & 17.1 & 14.7 & -38.6 & -61.0 & 66.1 $\to$ 44.3 & 64.4 $\to$ 26.8 & 76.1 $\to$ 5.5 \\
    PTv3 + block aug. & \textcolor{gray}{76.2} & 71.5 & 66.3 & 58.3 & 47.3 & 65.4 & 55.4 & 44.5 & 35.8 & 44.4 & 34.3 & 27.4 & 22.5 & -28.5 & -53.8 & 68.5 $\to$ 52.6 & 67.0 $\to$ 34.4 & 76.1 $\to$ 10.9 \\
    \method{} & \textcolor{gray}{71.5} & 71.2 & 69.6 & 67.8 & 64.0 & 69.7 & 67.1 & 63.8 & 59.2 & 64.2 & 60.0 & 56.7 & 53.8 & -7.6 & -17.7 & 61.5 $\to$ 54.0 & 56.8 $\to$ 44.4 & 71.5 $\to$ 50.2 \\
    \bottomrule
    \end{tabular}}
\end{table*}

\paragraph{ScanNet200.} \cref{tab:probe200} repeats the protocols of \cref{tab:probe} on ScanNet200~\cite{rozenberszki2022language}, whose $200$ classes make the probe far harder and reward the larger pretraining corpora of Concerto~\cite{zhang2025concerto} and Utonia~\cite{zhang2026utonia}, which lead ours by six to seven points on the clean scan. Under missing points the ordering of \cref{tab:probe} returns: ours loses $5.4$ mIoU on average and $10.9$ at worst over the six cells, against $7.5$ to $17.5$ on average and $12.9$ to $28.7$ at worst for the baselines, and from $70\%$ removed on it is the best model in absolute terms despite its lower starting point. On the objects that are almost entirely hidden, ours and Utonia share the first two places in every cell. Pretrained on the same two datasets, \method{}$^{\dagger}$ leads Sonata$^{\dagger}$~\cite{wu2025sonata} in every column.

\paragraph{ScanNet++.} \cref{tab:probe_snpp} repeats the protocols of \cref{tab:probe} on ScanNet++~\cite{yeshwanth2023scannet++}. As on ScanNet200, the larger pretraining corpora of Concerto~\cite{zhang2025concerto} and Utonia~\cite{zhang2026utonia} lead on the clean scan, here by about five points. Under missing points ours is the best model in every cell of the random-mask block, by $1.8$ to $3.2$ mIoU over the best baseline, and loses $7.3$ mIoU on average and $12.8$ at worst over the six cells, against $12.5$ to $15.6$ on average and $19.1$ to $22.3$ at worst for the baselines. On objects over $90\%$ hidden, ours is first in four of the five cells and second to Utonia in the fifth. Trained on the train splits only, \method{}$^{\dagger}$ leads Sonata$^{\dagger}$~\cite{wu2025sonata} in every cell, by $3.4$ on the clean scan and $10.8$ at $0.4$@$90$.

\paragraph{Outdoor Transfer.} \cref{tab:perclass_transfer} breaks \cref{tab:transfer} down by class. On both datasets, ours leads on most ground classes (driveable surface or road, sidewalk, other flat or other ground) and on cars, but trails the baselines on thin vertical structures (traffic cone; pole, trunk and traffic sign) and on people (pedestrian; bicyclist), and on nuScenes~\cite{caesar2020nuscenes} also on terrain, man-made structures and vegetation. Pretrained on the train splits of ScanNet and ScanNet++ alone, \method{}$^{\dagger}$ reaches $44.5$ on SemanticKITTI~\cite{behley2019semantickitti} and Sonata$^{\dagger}$~\cite{wu2025sonata} $38.8$.

\begin{table}[t]
    \centering
    \captionsetup{skip=3pt}
    \caption{Large holes vs.\ many small holes. ScanNet val mIoU on points within $0.1$\,m of a hole, with $70\%$ of the scene removed either as large $1.6$\,m holes or as many small $0.4$\,m holes. The scored points depend on the mask, so each mask has its own clean column, measured on the same points with the full scene as input. masked: mIoU and relative change from clean; perforated $-$ large: difference of the two relative changes, in percentage points. Both models are pretrained on the ScanNet and ScanNet++ train splits only.
    }
    \label{tab:strat}
    \setlength{\tabcolsep}{4pt}
    \small
    \resizebox{\linewidth}{!}{%
    \begin{tabular}{l cc cc c}
    \toprule
    \multirow{2}{*}{Model} & \multicolumn{2}{c}{large $1.6$\,m} & \multicolumn{2}{c}{perforated $0.4$\,m} & \multirow{2}{*}{\shortstack{perforated\\$-$ large}} \\
    \cmidrule(lr){2-3}\cmidrule(lr){4-5}
    & clean & masked & clean & masked & \\
    \midrule
    Sonata~\cite{wu2025sonata} & \textcolor{gray}{65.3} & \cellcolor{Salmon!22}55.8 (-14.6\%) & \textcolor{gray}{64.2} & \cellcolor{Salmon!30}49.6 (-22.7\%) & \cellcolor{Salmon!15}-8.1 \\
    \method{} & \textcolor{gray}{72.8} & \cellcolor{Salmon!18}64.8 (-10.9\%) & \textcolor{gray}{71.4} & \cellcolor{Salmon!11}67.5 (-5.5\%) & +5.4 \\
    \bottomrule
    \end{tabular}}
\end{table}

\paragraph{Supervised Models.} Missing points are not only a problem of self-supervised pretraining. \cref{tab:supervised} evaluates two fully supervised models, PTv3~\cite{wu2024point} and Volt-S~\cite{yilmaz2026volt}, under the protocol of \cref{tab:probe}. Both start about four points above ours on the complete scan, but fall below ours in every masked cell, already with the smallest holes; at $r{=}0.9$ they keep less than half of their clean performance, while ours keeps most of it.

\begin{table*}[t]
    \centering
    \captionsetup{skip=3pt}
    \caption{\textbf{Baselines of \cref{tab:explicit} with and without Seen2Scene~\cite{meng2026seen2scene}.} ScanNet val mIoU on the $31$ scenes of \cref{tab:explicit}, over all visible and removed points (left) and over the visible points alone (right), which the fill also changes; single view: the removed points of objects more than $75\%$ / $90\%$ hidden. w/o S2S: the encoder runs on the scan as observed; w/ S2S: on the scan completed by Seen2Scene, as in \cref{tab:explicit}. In both rows, each visible point reads its own feature; each removed point, scored only in the left block, reads the mean feature of its $8$ nearest points. The visible points are the same in both rows; the fill only adds generated points around them, so the right block isolates the damage the fill does to observed points.}
    \label{tab:explicit_controls}
    \tablestyle{3pt}{1.08}
    \resizebox{0.82\textwidth}{!}{%
    \begin{tabular}{l l | c c c | c c c | c c | c c c | c c c | c}
    \toprule
    & & \multicolumn{8}{c|}{\textit{Evaluated on visible + removed points}} & \multicolumn{7}{c}{\textit{Evaluated on visible points only}} \\
    \cmidrule(lr){3-10}\cmidrule(lr){11-17}
    \multirow{2}{*}{Encoder} & \multirow{2}{*}{Input} & \multicolumn{3}{c|}{$r{=}0.5$} & \multicolumn{3}{c|}{$r{=}0.7$} & \multicolumn{2}{c|}{Single View} & \multicolumn{3}{c|}{$r{=}0.5$} & \multicolumn{3}{c|}{$r{=}0.7$} & Single \\
    \cmidrule(lr){3-5}\cmidrule(lr){6-8}\cmidrule(lr){9-10}\cmidrule(lr){11-13}\cmidrule(lr){14-16}
    & & $b{=}0.2$ & $0.4$ & $0.8$ & $0.2$ & $0.4$ & $0.8$ & $>\!75\%$ & $>\!90\%$ & $b{=}0.2$ & $0.4$ & $0.8$ & $0.2$ & $0.4$ & $0.8$ & View \\
    \midrule
    MSM~\cite{hermosilla2025masked} & w/o S2S & 62.9 & 58.5 & 49.3 & 56.6 & 49.1 & 39.0 & 54.5 & 57.1 & 65.0 & 63.9 & 60.8 & 61.9 & 60.5 & 59.4 & 61.5 \\
     & w/ S2S & 26.6 & 29.0 & 29.5 & 15.8 & 16.7 & 16.5 & 29.9 & 32.2 & 29.6 & 35.4 & 42.8 & 18.5 & 24.2 & 31.3 & 34.7 \\
    \midrule
    Sonata~\cite{wu2025sonata} & w/o S2S & 65.5 & 59.5 & 52.0 & 56.1 & 43.5 & 39.7 & 57.5 & 60.1 & 68.1 & 66.4 & 63.8 & 61.5 & 55.2 & 62.0 & 65.0 \\
     & w/ S2S & 43.8 & 42.4 & 34.4 & 26.2 & 22.2 & 18.1 & 36.4 & 39.3 & 45.7 & 50.5 & 45.6 & 32.1 & 31.8 & 28.9 & 42.5 \\
    \midrule
    Concerto~\cite{zhang2025concerto} & w/o S2S & 71.5 & 65.5 & 57.5 & 63.6 & 53.3 & 46.1 & 66.7 & 68.5 & 73.9 & 72.2 & 70.2 & 69.3 & 66.3 & 69.7 & 74.5 \\
     & w/ S2S & 53.6 & 50.0 & 42.5 & 30.2 & 24.4 & 20.6 & 43.3 & 47.7 & 56.3 & 59.4 & 56.6 & 36.2 & 36.7 & 36.6 & 52.7 \\
    \midrule
    Utonia~\cite{zhang2026utonia} & w/o S2S & 71.5 & 66.0 & 58.9 & 65.4 & 57.2 & 47.6 & 67.8 & 70.1 & 73.5 & 71.9 & 70.8 & 70.8 & 69.0 & 72.1 & 74.7 \\
     & w/ S2S & 54.1 & 48.3 & 40.4 & 28.0 & 23.1 & 19.4 & 44.0 & 47.2 & 57.0 & 57.3 & 54.5 & 33.2 & 32.6 & 33.3 & 52.7 \\
    \bottomrule
    \end{tabular}}
\end{table*}

\begin{table}[t]
    \centering
    \captionsetup{skip=3pt}
    \caption{\textbf{Ours without oracle positions.} Whole-scene ScanNet val mIoU over the $31$ scenes of \cref{tab:explicit}, as in our row there. Oracle: the predictor is queried at the removed positions. Blind grid: it is queried at every node of a $2$\,cm lattice over the removed regions, and each removed point averages the features of its $8$ nearest nodes or visible points.}
    \label{tab:blind_grid}
    \tablestyle{2.5pt}{1.05}
    \resizebox{\linewidth}{!}{%
    \begin{tabular}{l | c c c | c c c | c c}
    \toprule
    \multirow{2}{*}{Query} & \multicolumn{3}{c|}{$r{=}0.5$} & \multicolumn{3}{c|}{$r{=}0.7$} & \multicolumn{2}{c}{Single View} \\
    & $b{=}0.2$ & $0.4$ & $0.8$ & $0.2$ & $0.4$ & $0.8$ & $>\!75\%$ & $>\!90\%$ \\
    \midrule
    Oracle & 67.7 & 66.7 & 62.6 & 65.2 & 63.1 & 54.5 & 61.0 & 62.8 \\
    Blind grid & 67.9 & 66.9 & 62.7 & 65.5 & 63.3 & 54.5 & 61.0 & 62.8 \\
    \bottomrule
    \end{tabular}}
\end{table}

\begin{table*}[t]
    \centering
    \captionsetup{skip=3pt}
    \caption{\textbf{Occupancy on the blind grid.} Grid nodes over the removed regions of the $31$ scenes of \cref{tab:explicit}. Rows: the occupancy probe of \cref{tab:occ}, thresholded on its surface probability $P$. Surface: a removed point within $2$\,cm, about $2\%$ of the nodes. Free space: no scan point within $5$\,cm, on the free side of the nearest surface; about half of the empty nodes, under a third in single view. Left: free-space nodes called surface, over all free-space nodes. Middle: surface nodes called empty, over all surface nodes. Right: free-space nodes called surface, over all nodes called surface; nodes in the $2$--$5$\,cm band and empty nodes inside solid objects enter only this denominator.}
    \label{tab:grid_occ}
    \tablestyle{2.5pt}{1.05}
    \resizebox{\textwidth}{!}{%
    \begin{tabular}{l | c c c c c c c | c c c c c c c | c c c c c c c }
    \toprule
    & \multicolumn{7}{c|}{\textit{Free space called surface} (\%) $\downarrow$} & \multicolumn{7}{c|}{\textit{Surface called empty} (\%) $\downarrow$} & \multicolumn{7}{c}{\textit{Fake share of claimed surface} (\%) $\downarrow$} \\
    \cmidrule(lr){2-8}\cmidrule(lr){9-15}\cmidrule(lr){16-22}
    & \multicolumn{3}{c}{$r{=}0.5$} & \multicolumn{3}{c}{$r{=}0.7$} & Single & \multicolumn{3}{c}{$r{=}0.5$} & \multicolumn{3}{c}{$r{=}0.7$} & Single & \multicolumn{3}{c}{$r{=}0.5$} & \multicolumn{3}{c}{$r{=}0.7$} & Single \\
    \cmidrule(lr){2-4}\cmidrule(lr){5-7}\cmidrule(lr){9-11}\cmidrule(lr){12-14}\cmidrule(lr){16-18}\cmidrule(lr){19-21}
    & $b{=}0.2$ & $0.4$ & $0.8$ & $0.2$ & $0.4$ & $0.8$ & View & $b{=}0.2$ & $0.4$ & $0.8$ & $0.2$ & $0.4$ & $0.8$ & View & $b{=}0.2$ & $0.4$ & $0.8$ & $0.2$ & $0.4$ & $0.8$ & View \\
    \midrule
    $P{>}0.5$ & 11.7 & 12.2 & 14.4 & 12.7 & 13.4 & 17.0 & 31.8 & 10.1 & 13.0 & 17.7 & 10.9 & 15.2 & 21.1 & 20.1 & 16.5 & 17.5 & 20.5 & 17.4 & 19.0 & 24.2 & 12.9 \\
    $P{>}0.7$ & 5.5 & 5.6 & 6.2 & 5.8 & 5.8 & 7.0 & 15.4 & 19.3 & 23.9 & 30.5 & 21.0 & 27.6 & 36.0 & 36.3 & 10.8 & 11.4 & 13.2 & 11.2 & 12.0 & 15.9 & 9.2 \\
    $P{>}0.9$ & 1.6 & 1.5 & 1.5 & 1.6 & 1.4 & 1.5 & 4.2 & 42.4 & 48.6 & 56.1 & 45.1 & 53.0 & 63.0 & 63.9 & 6.2 & 6.5 & 7.2 & 6.3 & 6.6 & 8.5 & 6.1 \\
    \bottomrule
    \end{tabular}}
\end{table*}

\paragraph{Is Augmentation Enough?} A natural fix for the supervised models is to train them on the same kind of input, which PTv3~\cite{wu2024point} retrained with block removal as an augmentation does (\cref{tab:aug}). Clean performance is unchanged, and inside the trained range the damage roughly halves, yet it remains at least five times ours. Outside that range the benefit fades with distance: the drop grows steadily with larger blocks and higher ratios, and at the largest holes it is about three times ours. On occlusion types the augmentation never produced, it helps little: the model still loses about half of its performance on objects over $90\%$ hidden under a single viewpoint and collapses under $99\%$ random dropout. Augmentation interpolates within what it has seen; it does not teach the model to reason about what is missing.

\paragraph{Many Small Holes or a Few Large Ones?} With the same fraction of the scene removed, many small holes should be easier to handle than a few large ones, since more context remains around each. \cref{tab:strat} tests this at $r{=}0.7$ on the points within $0.1$\,m of a hole, where the loss of context is felt most. Sonata~\cite{wu2025sonata}, pretrained on the same train splits, does the opposite: it loses $22.7\%$ with small holes against $14.6\%$ with large ones, and small holes are close to the masks it was pretrained with, where mask tokens told it where the geometry was. Ours behaves as the context argument predicts: small holes cost it half as much as large ones ($5.5\%$ vs.\ $10.9\%$).

\begin{figure*}[t]
    \centering
    \includegraphics[width=\linewidth]{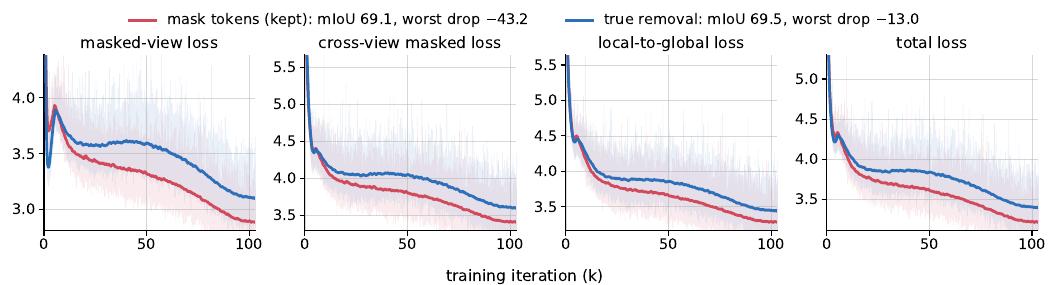}
    \caption{\textbf{A lower pretext loss does not mean better features.} Training losses of the mask-token and removal models of \cref{subtab:abl_removal}, which differ only in whether masked points are kept as tokens or removed. Once the mask schedule has warmed up, the mask-token model reaches a lower loss on every term: masked-view, cross-view masked, local-to-global and total. Yet it is far less robust: at the same clean performance, its drop under missing points is more than three times larger, on average and at worst (\cref{subtab:abl_removal}).}
    \label{fig:loss_token_vs_remove_grid}
\end{figure*}

\paragraph{Mask Tokens Make the Pretext Easier.} \cref{fig:loss_token_vs_remove_grid} compares the training losses of the first two models of \cref{tab:ablation-full}, which differ only in whether masked points are kept as tokens or removed. Once the mask schedule has warmed up, the token model reaches a lower loss on every term, yet it is the less robust of the two: the lower loss does not come from better features. The masked-view terms are easier because the tokens give away where the missing geometry is. The local-to-global term involves no masking, but its views pass through the same encoder and are matched to the same prototypes; a likely explanation is that an encoder allowed to rely on point positions settles on a simpler, position-driven prototype assignment that serves every view. This is the optimization shortcut of \cref{sec:sfc}: the pretext objective becomes easier to minimize, and robustness to missing points is what is lost.
\begin{table*}[t]
    \centering
    \caption{Full grid of the ablation in \cref{subtab:abl_removal}. ScanNet val
    mIoU with the protocol and fixed evaluation points of \cref{tab:probe}; all models pretrained on
    the ScanNet and ScanNet++ train splits with identical backbone, data and schedule.}
    \label{tab:ablation-full}
    \setlength{\tabcolsep}{3pt}
    \small
    \begin{tabular}{l c ccc ccc ccc}
    \toprule
    \multirow{2}{*}{Model} & \multirow{2}{*}{clean} & \multicolumn{3}{c}{$r=0.5$} & \multicolumn{3}{c}{$r=0.7$} & \multicolumn{3}{c}{$r=0.9$} \\
    \cmidrule(lr){3-5}\cmidrule(lr){6-8}\cmidrule(lr){9-11}
    & & $b{=}0.2$ & $0.4$ & $0.8$ & $0.2$ & $0.4$ & $0.8$ & $0.2$ & $0.4$ & $0.8$ \\
    \midrule
    Masked points kept as tokens & \textcolor{gray}{69.1} & 66.1 & 63.8 & 60.6 & 51.5 & 50.0 & 49.5 & 26.8 & 25.9 & 26.3 \\
    Masked points removed & \textcolor{gray}{69.5} & \cellcolor[HTML]{eef9f0}68.3 & \cellcolor[HTML]{eef9f0}67.0 & \cellcolor[HTML]{eef9f0}64.5 & \cellcolor[HTML]{eef9f0}66.6 & \cellcolor[HTML]{eef9f0}63.7 & \cellcolor[HTML]{eef9f0}60.4 & \cellcolor[HTML]{eef9f0}60.2 & \cellcolor[HTML]{eef9f0}56.5 & \cellcolor[HTML]{eef9f0}53.9 \\
    Removed + predictor (ours) & \textcolor{gray}{71.5} & \cellcolor[HTML]{def3e4}\bf 71.2 & \cellcolor[HTML]{def3e4}\bf 69.6 & \cellcolor[HTML]{def3e4}\bf 67.8 & \cellcolor[HTML]{def3e4}\bf 69.7 & \cellcolor[HTML]{def3e4}\bf 67.1 & \cellcolor[HTML]{def3e4}\bf 63.8 & \cellcolor[HTML]{def3e4}\bf 64.2 & \cellcolor[HTML]{def3e4}\bf 60.0 & \cellcolor[HTML]{def3e4}\bf 56.7 \\
    \bottomrule
    \end{tabular}
\end{table*}

\paragraph{Explicit Completion Baseline.} An explicit alternative to the predictor is to reconstruct the missing geometry and encode the completed scene. \cref{tab:explicit} tests this with Seen2Scene~\cite{meng2026seen2scene}, a recent generative scene completer, used zero-shot with its released weights, on $31$ ScanNet val scenes, since completion takes minutes per scene. The cell mask is applied to the RGB-D frames, which are fused with the holes carved out; the completer fills the holes without its layout boxes, which would reveal object positions, and the vertices of the generated surface inside the removed regions join the visible points at the encoders' $2$\,cm grid, with colours from the nearest visible point and normals from the generated mesh. Each encoder runs on the completed scene, reads visible points from their own features, and reads each removed point from the averaged features of its $k$ nearest points, generated or visible, since the generated surface does not pass through the removed points exactly. Ours runs on the unfilled scan and reads every removed position from the predictor. We score the whole scene, so that damage a wrong fill does to the visible points counts as well; in the single-view columns, the removed points are those of the objects more than $75\%$ / $90\%$ hidden.

Filling the holes lowers every baseline below its unfilled scan in every setting (\cref{tab:explicit_controls}), by $15$ to $41$ points over the whole scene and $13$ to $43$ on the visible points alone. The generated surfaces cover about half of the removed points at $0.2$\,m holes and a quarter at $0.8$\,m, and most of what they add lies where no point was removed; this geometry enters the context of the visible points and degrades them.
Neither a seam-free sampling mode nor thinning the fill to the density of the scan closes the gap. The completer was trained on partial scans made by dropping whole frames, which leaves large unobserved regions rather than many small holes, but under the single-view mask, the occlusion closest to that setting, it is still the worst input for every encoder. It is also far slower: more than $20$ minutes of completion per scene on one A100.

\paragraph{Without Oracle Positions.} In \cref{tab:explicit}, our predictor queries the removed positions themselves, which a deployed agent may not know. We therefore also query a blind grid: a $2$\,cm lattice over the removed regions, every node read by the predictor, and each scored point averages the features of its $8$ nearest nodes or visible points, as in the $k$NN readout of \cref{tab:amodal}. The score stays within $0.3$ of the oracle positions in every column of \cref{tab:blind_grid}, so the true positions only decide where the features are read, not how good they are. Like Seen2Scene~\cite{meng2026seen2scene}, which receives the unobserved space as part of its fused input, the grid knows where the holes are, but not what lies in them. 

\paragraph{Occupancy on the Grid.} Semantics are scored at real points only, so a grid node that claims a surface in an empty space costs nothing above. We measure this separately in \cref{tab:grid_occ}: the occupancy probe of \cref{tab:occ} labels every grid node as surface or empty at three thresholds on its surface probability. A node is a true surface if a removed point lies within $2$\,cm, and empty only if no scan point lies within $5$\,cm; nodes in between are left out, since sensor noise and grid spacing make them ambiguous. Of the empty nodes, we count only those on the free side of the nearest surface, decided by a vote of the normals of the nearest scan points: calling the inside of a solid object occupied does not mislead an agent, whereas a false surface in free space, under a table or in front of a sofa, does. We report how much free space is called surface, how much true surface is called empty, and how much of the surface the probe claims lies in free space. A high threshold calls almost no free space surface and keeps nearly all of its claims out of free space, but misses much of the hidden surface; a low threshold finds most of the surface while calling more free space surface. The threshold thus lets an agent choose between caution and coverage. In single view, where whole objects are hidden, the probe calls the most free space surface.

\section{More Visualization}
\label{sec:more_vis}

\begin{figure*}[t]
    \centering
    \includegraphics[width=1\textwidth]{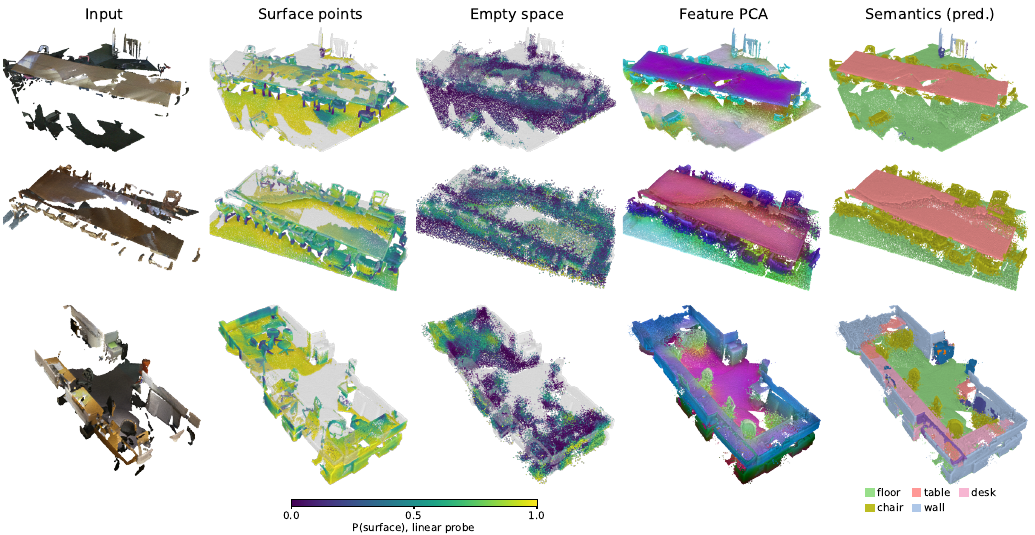}
    \caption{\textbf{More examples of predicting an unobserved region} (1/2).}
    \label{fig:occ_field_supp1}
\end{figure*}

\begin{figure*}[t]
    \centering
    \includegraphics[width=1\textwidth]{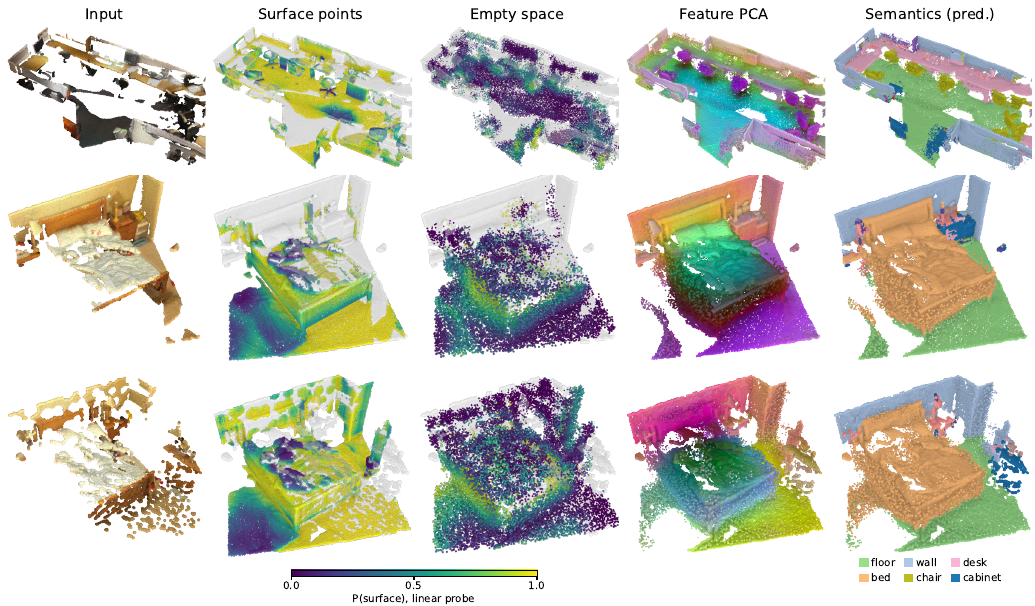}
    \caption{\textbf{More examples of predicting an unobserved region} (2/2).}
    \label{fig:occ_field_supp2}
\end{figure*}

\paragraph{Predicting Unobserved Regions.} \cref{fig:occ_field_supp1,fig:occ_field_supp2} show more scenes in the format of \cref{fig:occ_field}. 
Large objects with a variable extent, such as beds and tables, make the boundary between surface and empty space hardest to place from the visible part alone.

\clearpage

%%%%%%%%%%%%%%%%%%%%%%%%%%%%%%%%%%%%%%%%%%%%%%%%%%%%%%%%%%%%

\end{document}